\documentclass{article}

\usepackage{arxiv}

\usepackage[utf8]{inputenc} 
\usepackage[T1]{fontenc}    
\usepackage{hyperref}       
\usepackage{url}            
\usepackage{booktabs}       
\usepackage{amsfonts}       
\usepackage{nicefrac}       
\usepackage{microtype}      
\usepackage{lipsum}

\usepackage{tikz}
\usepackage{amsmath, amssymb, latexsym}
\usepackage{multirow}
\usepackage{breqn}
\usepackage{labelschanged}

\usetikzlibrary{positioning}

\title{Generative adversarial networks in time series: A survey and taxonomy}

\author{
  \makebox[.4\linewidth]{Eoin Brophy} \\
  Infant Research Centre \& School of Computing\\
  Dublin City University\\
  Ireland\\
  \texttt{eoin.brophy7@mail.dcu.ie} \\
    \And
 \makebox[.4\linewidth]{Zhengwei Wang}\\
  ByteDance AI Lab\\
  China\\
   \And
 \makebox[.4\linewidth]{Qi She}\\
  ByteDance AI Lab\\
  China\\
 \\
    \And
 \makebox[.4\linewidth]{Tom{\'a}s Ward}\\
  Insight SFI Research Centre for Data Analytics\\
  Dublin City University\\
  Ireland \\
}

\begin{document}
\maketitle
\begin{abstract}
  Generative adversarial networks (GANs) studies have grown exponentially in the past few years. Their impact has been seen mainly in the computer vision field with realistic image and video manipulation, especially generation, making significant advancements. While these computer vision advances have garnered much attention, GAN applications have diversified across disciplines such as time series and sequence generation. As a relatively new niche for GANs, fieldwork is ongoing to develop high quality, diverse and private time series data. In this paper, we review GAN variants designed for time series related applications. We propose a taxonomy of discrete-variant GANs and continuous-variant GANs, in which GANs deal with discrete time series and continuous time series data. Here we showcase the latest and most popular literature in this field; their architectures, results, and applications. We also provide a list of the most popular evaluation metrics and their suitability across applications. Also presented is a discussion of privacy measures for these GANs and further protections and directions for dealing with sensitive data. We aim to frame clearly and concisely the latest and state-of-the-art research in this area and their applications to real-world technologies.
\end{abstract}

\keywords{Generative Adversarial Networks \and Time Series \and Discrete-variant GANs \and Continuous-variant GANs}

\section{Introduction}
\label{sec:intro}
This review paper is designed for those interested in GANs applied to time series data generation. We provide a review of current state-of-the-art and novel time series GANs and their solutions to real-world problems. The applicability of GANs to this type of data can solve many issues that current dataset holders face. Data shortage is often an issue, and GANs can augment smaller datasets by generating new, previously unseen data. Data can be missing or corrupted in cases; GANs can impute data, i.e. replace the artefacts with information representative of clean data. GANs are also capable of denoising signals in the case of corrupted data. Data protection, privacy, and sharing have become heavily regulated; GANs can ensure an extra layer of data protection by generating differentially private datasets containing no risk of linkage from source to generated datasets. 

\begin{figure}[ht]
    \centering
    \includegraphics[width=\columnwidth]{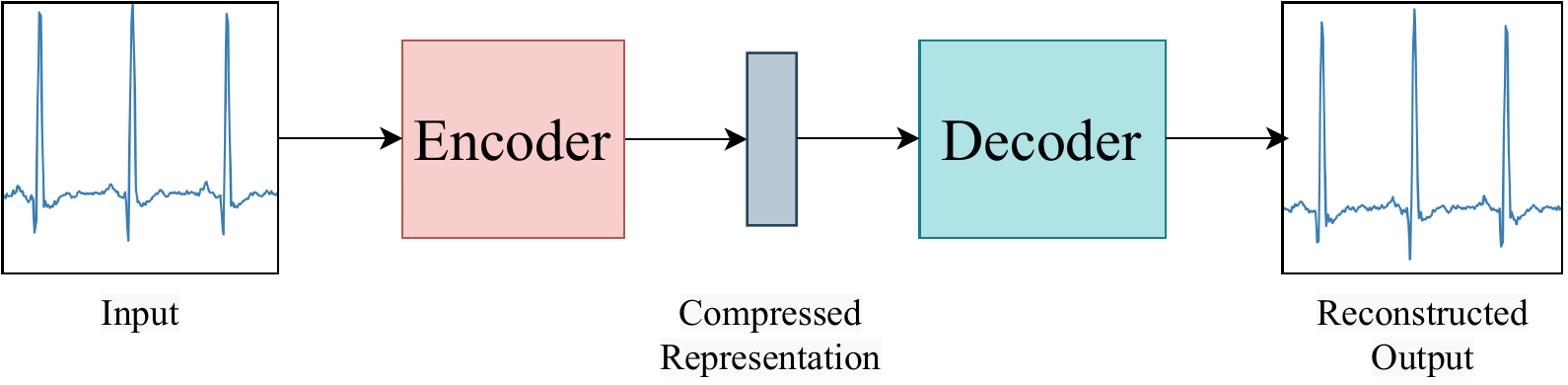}
    \caption{Autoencoder Model}
    \label{fig:AE_model}
\end{figure}

Several methods have been used in the past to generate synthetic data. One such method is the autoencoder (AE) which is designed to efficiently learn an informative representation of an input in a small dimensional space and reconstruct the encoded data back such that the reconstructed input is similar as possible to the original one. The AE model is made of an encoder and decoder neural network, as shown in Figure \ref{fig:AE_model}. However, other generative models have emerged as front-runners due to the quality of the generated data and inherent privacy protection measures.

Generative Adversarial Networks have been gaining a lot of traction amongst the deep learning research community since their inception in 2014 \cite{Goodfellow2014}. Their ability to generate and manipulate data across multiple domains has contributed to their success. While the main focus of GANs to date has been in the computer vision (CV) domain, they have also been successfully applied to others, such as natural language processing (NLP). There has also been a movement towards the use of GANs for time series and sequential data generation, and forecasting.

A GAN is a generative model consisting of a generator and discriminator, typically two neural networks (NN) models. In recent years GANs have demonstrated their ability to produce high-quality image and video generation, style-transfer, and image completion. They have also been successfully used for audio generation, sequence forecasting, and imputation.

However, one of the significant challenges of GANs lies in their inherent instability, which makes it difficult to train. GAN models suffer from issues such as non-convergence, diminishing/vanishing gradients, and mode collapse. A non-converging model does not stabilise and continuously oscillates, causing it to diverge. Diminishing gradients prevents the generator from learning anything as the discriminator becomes too successful. Mode collapse is when the generator collapses, producing only uniform samples with little to no variety.

The second challenge of GANs lies in its evaluation process. With image-based GANs, researchers have reached a loose consensus \cite{borji_pros_2018} surrounding the evaluation of the generated distribution estimated from the training data distribution. Unfortunately for time series GANs, due to the comparatively low numbers of papers published, there has not been an agreement reached on the generated data's evaluation metrics. There have been different approaches put forward, but none established as a front runner in the metrics space as of yet.


We define a time series as a sequence of vectors dependant on time $(t)$ and can be represented as $xt = {x1,..., xn}$ for continuous/real-time and discrete-time. The time series' values can either be defined as continuous or discrete and, depending on the number of values recorded, are univariate or multivariate. In most cases, the time series will take either an integer value or a real value. 

As Dorffner states, a time series can be viewed, from a practical perspective, as a value sampled at discrete steps in time \cite{Dorffner96neuralnetworks}. This time-step can be as long as years to as short as milliseconds, for example. We define a continuous time series as a signal sampled from a continuous process, i.e. the function's domain is from an uncountable set. In contrast, a discrete time series has a countable domain. 


In this review, we present the first complete review and taxonomy of time series GANs, namely discrete and continuous variants, their applications, architecture, loss functions and how they have improved on their predecessors in terms of variety and quality of their generated data. We also contribute by including experiments for the majority of time series GAN architectures applied to time series synthesis.



\section{Related Work}
\label{sec:reviews}
There has been a handful of high-quality GAN review papers published in the past few years. For example, Wang \textit{et al.} takes a taxonomic approach to GANs in Computer Vision \cite{Wang2020}. The authors split GANs into architecture variants and loss variants. While they include applications of GANs and mention their applicability to sequential data generation, the work is heavily focused on media manipulation and generation. The authors in \cite{Gui2020} breakdown GANs into their constituent parts. They begin by discussing the algorithms and architecture of various GANs and their evaluation metrics, then list their surrounding theory and problems such as mode collapse, amongst others. Finally, they discuss the applications of GANs and provide a very brief account of GANs used for sequential data. Gonog and Zhou \cite{Gonog2019} provide a short introduction to GANs, their theory and explores the variety of plausible models, again listing their applications in image and video manipulation with a mention of sequential data (NLP). In another review, paper \cite{Alqahtani2019} the authors give an overview of GAN fundamentals, variants, and applications. Sequential data applications are mentioned in the form of music and speech synthesis.

As with most review papers, Yinka-Banjo and Ugot give an introduction and overview of Generative Adversarial Networks \cite{Banjo2020}. However, they also review GANs as adversarial detectors and discuss their limitations applied to cybersecurity. Yi, Walia, and Babyn \cite{Yi2019} give a review of GANs and their applications in medical imaging, how they can be used in clinical research and potentially deployed to help practising clinicians. There is no mention of time series data use cases.

A recurring theme in these papers focuses on GAN variants which have mostly been applied to the computer vision domain. To the best of our knowledge, no review paper has been conducted with the main focus on time series GANs. While these reviews have mentioned the application of these GANs in generating sequential data, they have scratched the surface of what is becoming a growing body of research. 

We contribute to lessening this gap by presenting our work which is concerned with presenting the latest up-to-date research around time series GANs, their architecture, loss functions, evaluation metrics, trade-offs and approaches to privacy preservation of their datasets.


\section{Generative Adversarial Networks}
\label{sec:gan}
\subsection{Background}
\begin{figure}[ht]
    \centering
    \includegraphics[width=\columnwidth]{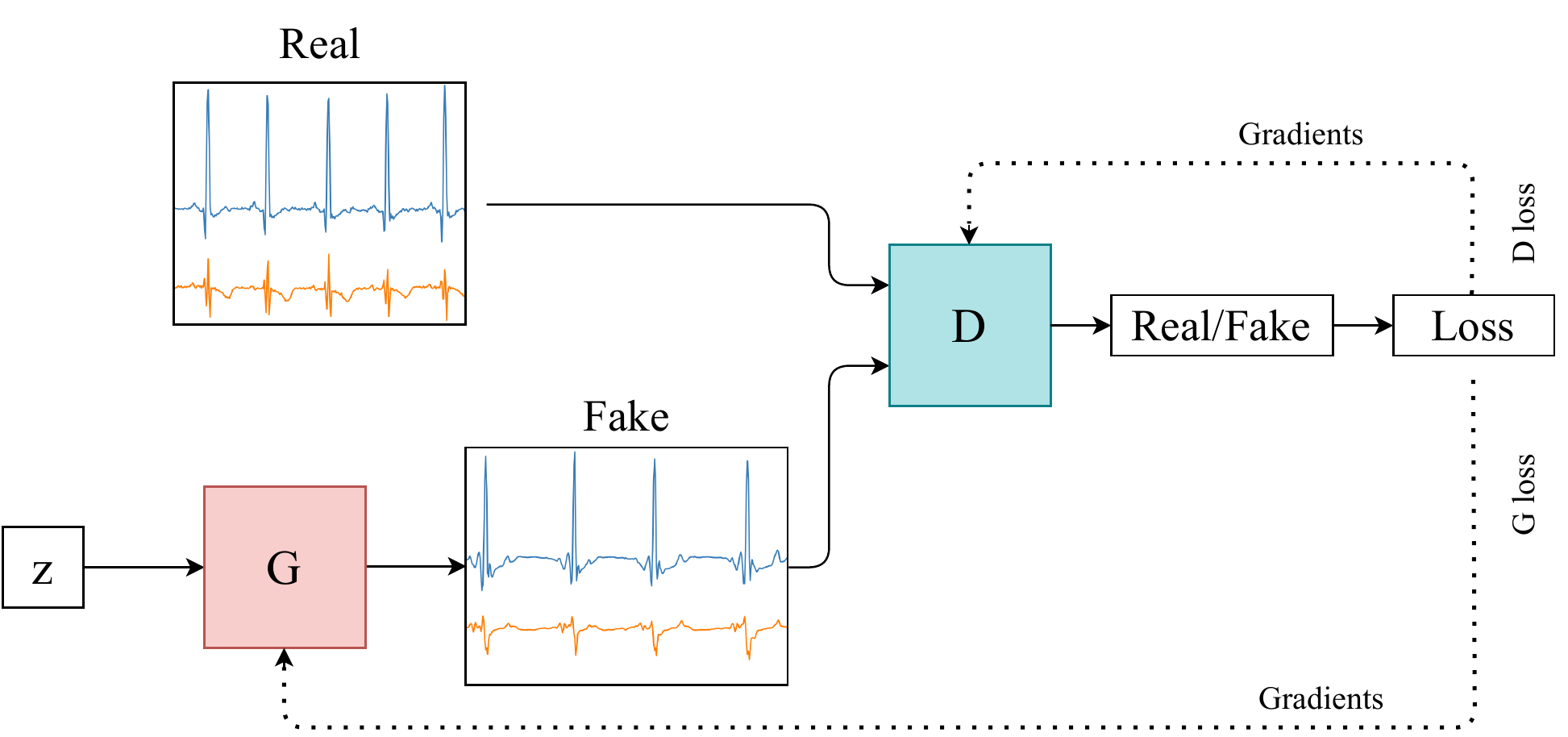}
    \caption{Generative Adversarial Network}
    \label{fig:GAN_model}
\end{figure}

The introduction of GANs facilitated a significant breakthrough in the generation of synthetic data. These deep learning models typically consist of two neural networks, a generator and a discriminator. The generator \textit{G} takes in random noise $\textbf{\textit{z}} \in \mathbb{R}^{r}$ and attempts generates synthetic data that is similar to the training data distribution. The discriminator \textit{D} attempts to determine if the generated data is real or fake. The generator aims to maximise the failure rate of the discriminator, while the discriminator aims to minimise it. Figure \ref{fig:GAN_model} shows a simple example of the GAN architecture and the game that the neural network models play. The two networks are locked in a two-player minimax game defined by the value function \textit{V(G,D)} (\ref{eq:1}), where \textit{D(\textbf{x})} is the probability that \textit{x} comes from the real data rather than the generated data \cite{Goodfellow2014}.

\begin{equation}
\mathop{min}_{G} \mathop{max}_{D}V(G,D)= \mathbb{E}_{x \sim p_{data}(x)}[logD(\textbf{x})] + \mathbb{E}_{z \sim p_{\textbf{z}}(z)}[log(1-D(G(\textbf{z})))]
\label{eq:1}
\end{equation}

GANs belong to the family of generative models and are an alternative method of generating synthetic data that do not require domain expertise. They were conceived in the paper by Goodfellow in 2014, where a multi-layer perceptron was used for both the discriminator and the generator \cite{Goodfellow2014}. Radford \textit{et al.} (2015) subsequently developed the deep convolutional generative adversarial network (DCGAN) to generate synthetic images \cite{Radford2015}. Since then, researchers have continuously improved on the early GAN architectures, loss functions, and evaluation metrics while innovating on their potential contributions to real-world applications. To appreciate why there has been such concerted activity in the further development of GAN technologies it is important to understand the limitations of early architectures and the challenges these presented. We describe these next, and in so doing, prepare the reader for the particular manifestation of these challenges in the more specific context of time series.

\subsection{Challenges}
\label{sec:challenges}
There are three main challenges in the area of time series GANs, i.e., training stability, evaluation and privacy risk associated with synthetic data created by GANs. We are going to explain these three challenges as follows.

\textbf{Training stability.}\hspace{5pt} The original work~\cite{Goodfellow2014} has already proved the global optimality and the convergence of GANs during training. It still highlights the instability problem that can arise when training a GAN. Two problems are well-studied in the literature 1. vanishing gradients and 2. mode collapse. The vanishing gradient is caused by directly optimizing loss presented in equation~\eqref{eq:1}. When $D$ reaches the optimality, optimizing the equation~\eqref{eq:1} for $G$ can be converted to minimising the Jensen-Shannon (JS) divergence (details of derivation can refer to section 5 in~\cite{Wang2020}) between $p_r$ and $p_g$:
\begin{equation}
    \mathcal{L}_{G}=2\cdot JS(p_{r}\|p_{g}) - 2 \cdot\mathrm{log}2
    \label{eq:JS-loss-G}
\end{equation}
JS divergence stays constant ($log2=0.693$) when there is no overlap between $p_r$ and $p_g$, which indicates the gradient for $G$ using this loss is 0 in this situation. Non-zero gradient for $G$ only exists when $p_r$ and $p_g$ have substantial overlap. In practice, the possibility that $p_r$ and $p_g$ are not intersected or have negligible overlap is very high~\cite{Arjovsky2017}. In order to get rid of the vanishing gradient problem for $G$, the original GAN work~\cite{Goodfellow2014} highlights that the minimization of
\begin{equation}
    \mathcal{L}_{G}=-\mathbb{E}_{\mathbf{x}\sim p_{g}}\mathrm{log}[D(\mathbf{x})]
    \label{eq:KL-loss-G}
\end{equation}
for updating $G$. This strategy is able to avoid the vanishing gradient problem but lead to the mode collapse issue. Optimizing equation~\eqref{eq:KL-loss-G} can be converted to optimizing the reverse Kullback–Leibler (KL) divergence i.e., $KL(p_g||p_r)$ (details can also refer to~\cite{Wang2020}). When $p_r$ contains multiple modes, $p_g$ chooses to recover a single mode and ignores other modes when optimizing the reverse KL divergence. Considering this case, $G$ trained using equation~\eqref{eq:KL-loss-G} might be only able to generate few modes from real data. These problems can be amended by changing the architecture or the loss function, which are reviewed by Wang~\textit{et al.}~\cite{Wang2020} in detail.

\textbf{Evaluation.}\hspace{5pt} A wide range of evaluation metrics has been proposed to evaluate the performance of GANs~\cite{Borji2019,borji2021pros,wang2020synthetic,wang2020use}. Current evaluations of GANs in computer vision are normally designed to consider two perspectives i.e., quality and quantity of generated data. The most representative qualitative metric is to use human annotation to determine the visual quality of the generated images. Quantitative metrics compare statistical properties between generated and real images i.e., two-sample tests such as maximum mean discrepancy (MMD)~\cite{Sutherland2016}, Inception Score~\cite{salimans_IS_2016} and Fr\'{e}chet Inception Distance (FID)~\cite{heusel2018FID}. Contrary to evaluating image-based GANs, it is difficult to evaluate time series data from human psycho-perceptual sense qualitatively. In terms of qualitatively evaluating time series based GANs, it normally conducts
t-SNE~\cite{tSNE_2008} and PCA~\cite{bryant1995principal} analyses to visualize how well the generated distributions resemble the original distributions~\cite{yoon_timeseriesGAN_2019}. Quantitative evaluation for time series based GANs can be done by deploying two-sample tests similar to image-based GANs.

\textbf{Privacy risk.}\hspace{5pt} Apart from evaluating the performance of GANs, a wide range of methods have been used to asses the privacy risk associated with synthetic data created by GANs. Choi \textit{et al.} performed tests for presence disclosure and attribute disclosure. In contrast, others utilised a three-sample test on the training, test, and synthetic data to identify if the synthetic data has overfitted to the training data \cite{Choi2017, Esteban2017}. It has been shown that common methods of de-identifying data do not prevent attackers from re-identifying individuals using additional data \cite{ElEmam2011,Malin2001}. Sensitive data is usually de-identified by removing personally identifiable information (PII). However, work is ongoing to create frameworks to link different sources of publicly available information together using alternative information to PII. Malin \textit{et al.} developed a software program, REID, to connect individuals contained in publicly available hospital discharge data with their unique DNA records \cite{Malin2001}. Culnane \textit{et al.} re-identified individuals in a de-identified open dataset of Australian medical billing records using unencrypted parts of the records and known information about individuals from other sources \cite{Culnane2017}. Hejblum \textit{et al.} developed a probabilistic method to link de-identified EHR data of patients with rheumatoid arthritis \cite{Hejblum2019}. The re-identification of individuals in publicly available datasets can lead to the exposure of their sensitive health information. Health data has been categorised as special personal data by General Data Protection Regulation (GDPR) and is subject to a higher level of protection under the Data Protection Act 2018 (Section36(2)) \cite{GDPR36_2}. Consequently, concerned researchers must find alternative methods of protecting sensitive health data to minimise the risk of re-identification. This will be addressed in Section \ref{sec:privacy}.


\subsection{Popular Datasets}
Unlike image-based datasets (CIFAR, MNIST, ImageNet \cite{Krizhevsky09_CIFAR, lecun_mnisthandwrittendigit_2010,imagenet_cvpr09}) there are no standardised or commonly used benchmarking datasets for time series generation. However, we have compiled a list of some of the more popular datasets implemented in the reviewed works, and they are listed in Table \ref{table:Popular Datasets}. There exist two repositories; the UCR Time Series Classification/Clustering database \cite{UCRArchive2018}, and the UCI Machine Learning repository \cite{UCIArchive2017} that make available several time series datasets. Despite this, there is still no consensus on a standardised dataset used for benchmarking time series GANs, which may be due to the `continuous' nature of the architecture dimensions. GANs designed for continuous time series generation often differ in the length of their input sequence due to either author preference or the constraints placed on their architecture for the generated data's downstream tasks.

\begin{table}[ht]
    \centering
    \caption{Popular Datasets used in the reviewed works.}
    \label{table:Popular Datasets}
    \begin{tabular}{  p{0.35\columnwidth}  p{0.2\columnwidth}  p{0.1\columnwidth} p{0.1\columnwidth} }
        \toprule
\textbf{Name (Year)}      
& \textbf{Data Type}   
& \textbf{Instances}
& \textbf{Attributes} \\\midrule

Oxford-Man Institute "realised library" (Updated Daily)
& Real Multivariate Time Series
& >2,689,487
& 5\\\hline

EEG Motor Movement/Imagery Dataset (2004)
& Real Multivariate Time Series 
& 1,500
& 64 \\\hline

ECG 200 (2001)
& Real Univariate Time Series 
& 200
& 1 \\\hline

Epileptic Seizure Recognition Dataset (2001)
& Real Multivariate Time Series 
& 11,500
& 179 \\\hline

TwoLeadECG (2015)
& Real Multivariate Time Series 
& 1,162
& 2 \\\hline

MIMIC-III (-)
& Real, Integer \& Categorical Multivariate Time Series 
& -
& - \\\hline

EPILEPSIAE project database (-)
& Real Multivariate Time Series 
& 30
& - \\\hline

PhysioNet/CinC (2015)
& Real Multivariate Time Series 
& 750
& 4 \\\hline

Wrist PPG During Exercise (2017)
& Real Multivariate Time Series 
& 19
& 14 \\\hline

MIT-BIH Arrhythmia Database (2001)
& Real Multivariate Time Series 
& 201
& 2 \\\hline

PhysioNet/CinC (2012)
& Real, Integer \& Categorical Multivariate Time Series 
& 12000
& 43 \\\hline

KDD Cup Dataset (2018)
& Real, Integer \& Categorical Multivariate Time Series 
& 282
& 3 \\\hline

PeMS Database (Updated Daily)
& Integer \& Categorical Multivariate Time Series 
& -
& 8 \\\hline

Nottingham Music Database (2003)
& Special Text Format Time Series 
& 1000
& - \\
        \bottomrule
    \end{tabular}
\end{table}

\section{Taxonomy of Time Series based GANs}
\label{sec:time-series-gans}

\begin{figure}[ht]
    \centering
    \includegraphics[width=0.49\columnwidth]{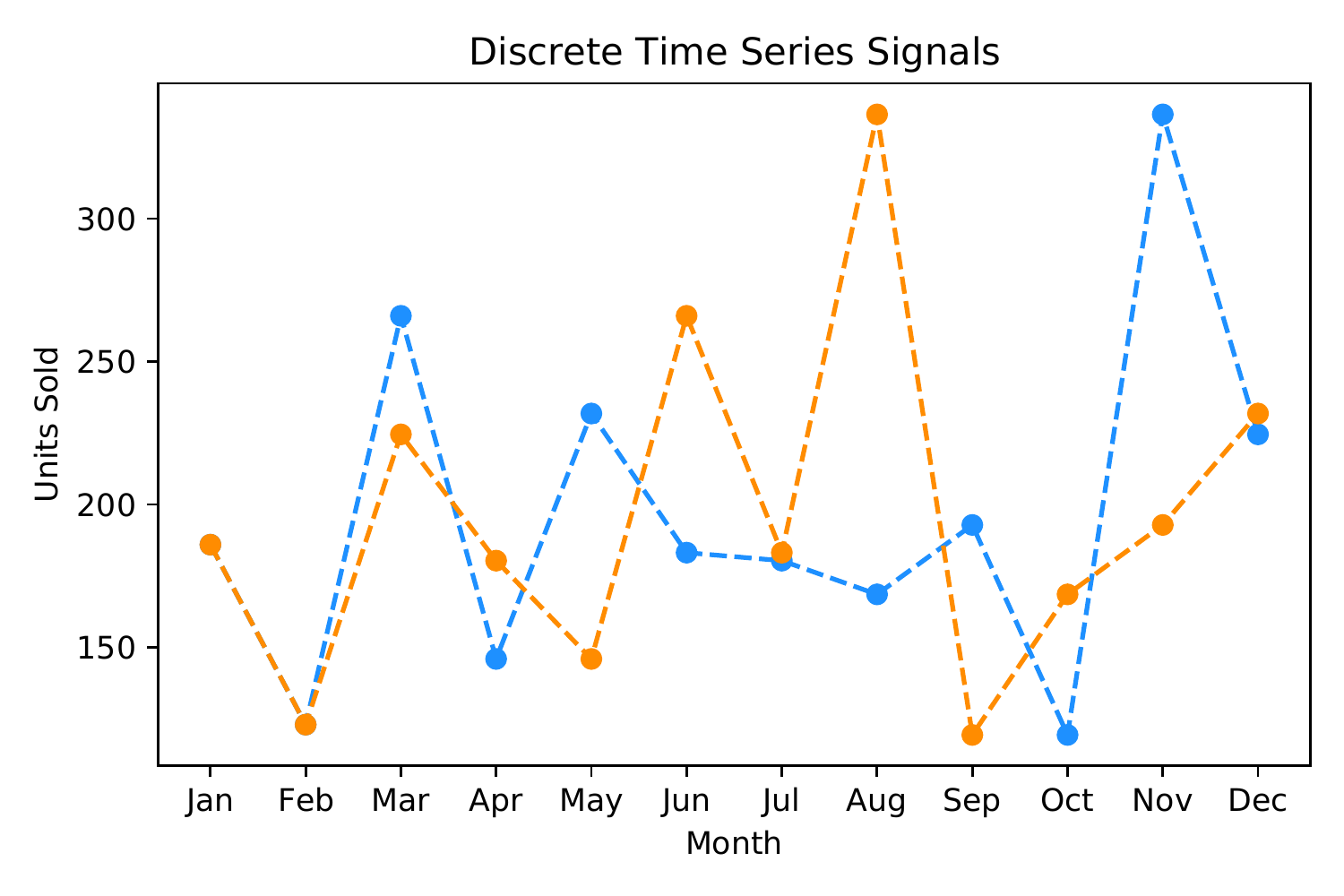}
    \includegraphics[width=0.49\columnwidth]{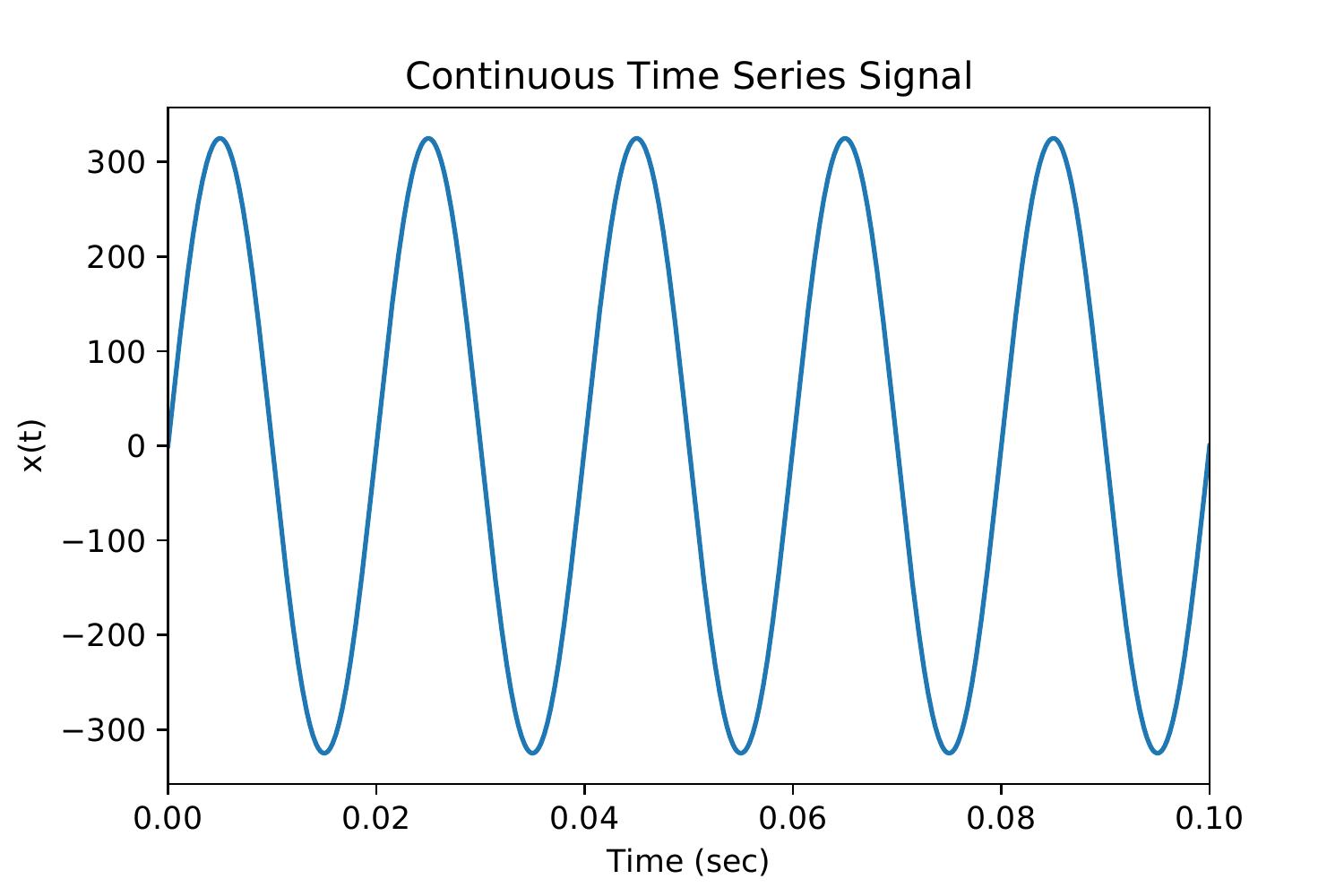}
    \caption{Example plots of discrete (left) and continuous time series (right).}
    \label{fig:Time_Series_Types}
\end{figure}

We propose a taxonomy of the following time series based GANs based on two distinct variant types: \textbf{discrete variants} (discrete time series) and \textbf{continuous variants} (continuous time series). A discrete time series consists of data points separated by time intervals. This type of data might have 1. a data-reporting interval that is infrequent (e.g., 1 point per minute) or irregular (e.g., whenever a user logs in) and 2. gaps where values are missing due to reporting interruptions (e.g., intermittent server or network downtime in a network traffic application). Discrete time series generation involves generating sequences that may have a temporal dependency but contain discrete tokens; these can be commonly found in electronic health records (International Classification of Diseases 9 codes) and text generation. A continuous time series has a data value corresponding to every moment in time. Continuous data generation is concerned with generating a real-valued signal x with temporal dependencies where x $\in \mathbb{R}$. See Figure \ref{fig:Time_Series_Types} for examples of discrete and continuous time series signals.

\textbf{Challenges with discrete time series generation.}\hspace{5pt} GANs struggle with discrete data generation due to the zero gradient nearly everywhere, i.e., the distribution on discrete objects are not differentiable with respect to their parameters \cite{Yu_seqgan_2017, hjelm2018boundaryseeking}.  This limitation makes the generator untrainable using backpropagation alone. The generator starts with a random sampling and a deterministic transform guided via the gradient of the loss from the discriminator with respect to the output produced by $G$ and the training dataset. This loss leads to a slight change in $G$'s output, pushing it closer to the desired output. Making slight changes to continuous numbers make sense; adding 0.001 to a value of 10 in financial time series data will bring it to 10.001. However, a discrete token such as the word ‘penguin’ cannot simply undergo the addition of 0.001 as the sum ‘penguin+0.001’ makes no sense.  What's important here is the impossibility for the generator to jump from one discrete token to the next because the small change gives the token a new value that does not correspond to any other token over that limited discrete space \cite{GoodfellowReddit2016}. This is because there exists 0 probability in the space between these tokens, unlike with continuous data. 

\textbf{Challenges with continuous time series generation.}\hspace{5pt} Modelling continuous time series data presents a different problem for GANs, which are inherently designed to model continuous data, albeit most commonly in the form of images. The temporal nature of continuous data in time series presents an extra layer of difficulty. Complex correlations exist between the temporal features and their attributes, e.g., if using multichannel biometric/physiological data, the ECG characteristics will depend on the individual's age and/or health. Also, long-term correlations exist in the data, which are not necessarily fixed in dimension compared to image-based data under a fixed dimension.
Transforming image dimensions may lead to a degradation in image quality, but it is a recognised practice. This operation becomes more difficult with continuous time series data as there is no standardised dimension used across time series GANs architectures, which means that benchmarking their performances becomes difficult.

Since their inception in 2014, GANs have shown great success in generating high-quality synthetic images indistinguishable from real images \cite{Guibas2017,Ledig2017,Reed2016}. While the focus to date has been on developing GANs for improved media generation, there is a growing consensus that GANs can be used for more than image generation and manipulation, which has led to a movement towards generating time series data with GANs. 

\begin{figure}[ht]
    \centering
    \includegraphics[width=0.85\columnwidth]{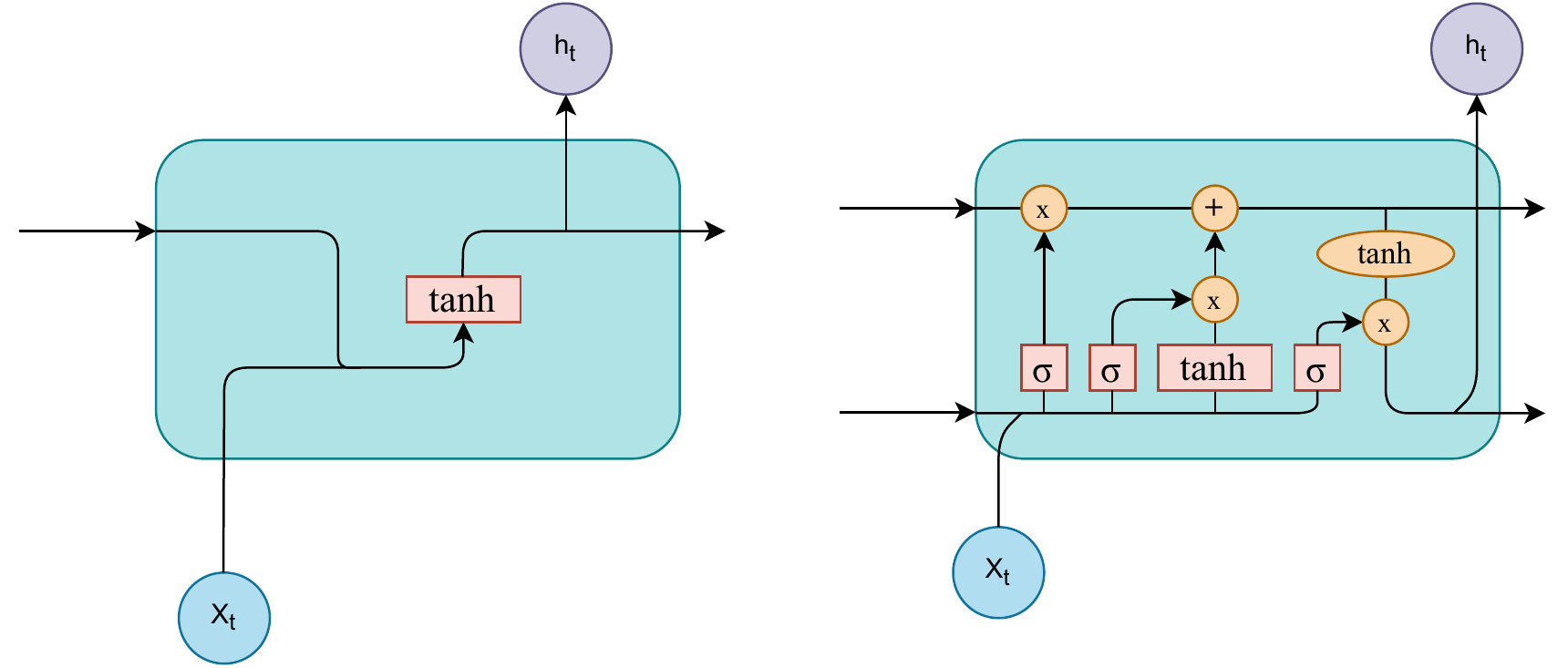}
    \caption{Block Diagram of (left) a standard RNN and (right) LSTM cell}
    \label{fig:SimpleRNN}
\end{figure}

Recurrent neural networks (RNNs) (Figure \ref{fig:SimpleRNN}), due to their loop-like structure, are perfect for sequential data applications but by themselves lack the ability to learn long-term dependencies that might be crucial in forecasting future values based on past. Long-short Term Memory networks (LSTM) (Figure \ref{fig:SimpleRNN} are a specific kind of RNN that have the ability to remember information for long periods of time and, in turn, learn these long-term dependencies that the standard RNN is not capable of doing. In most work reviewed in this paper, the majority of the RNN based architectures are utilising the LSTM cell.

RNNs can model sequential data such as financial data, medical data, text, and speech, and they have been the foundational architecture for time series GANs. A recurrent GAN (RGAN) was first proposed in 2016. The generator contained a recurrent feedback loop that used both the input and hidden states at each time step to generate the final output \cite{Im2016}. Recurrent GANs often utilise Long Short-Term Memory neural networks in their generative models to avoid the vanishing gradient problem associated with more traditional recurrent networks \cite{Hochreiter1997}. In the section that follows, we chronologically present time series GANs that have either contributed significantly to this space or have made some of the most recent novel advancements.

\subsection{Discrete-variant GANs}
\subsubsection{Sequence GAN (SeqGAN) (Sept. 2016)}
\begin{figure}[ht]
    \centering
    \includegraphics[width=0.85\columnwidth]{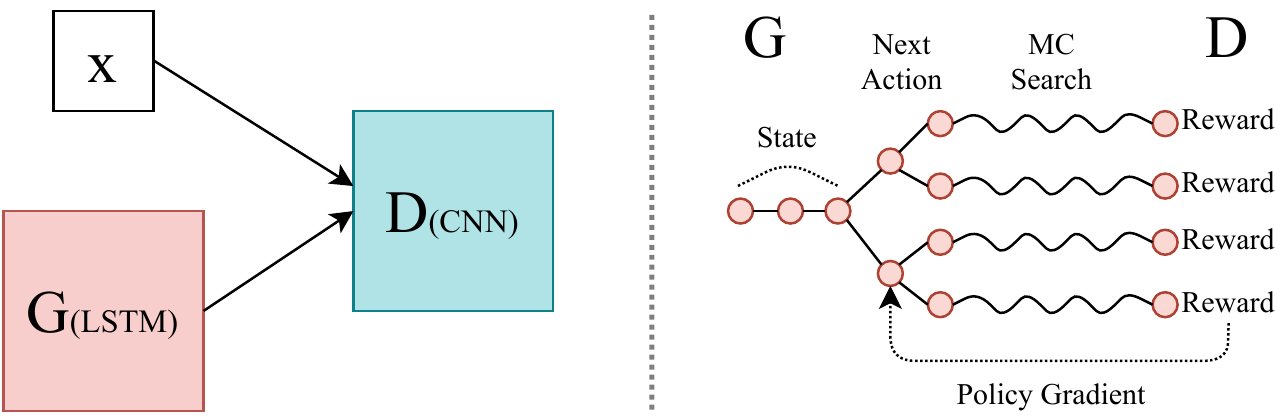}
    \caption{SeqGAN: \textit{D} (left) is trained over real and generated data, whereas \textit{G} (right) is trained by policy gradient where the final reward signal is provided by \textit{D} and is passed back to the intermediate action value via Monte Carlo search \cite{Yu_seqgan_2017}.}
    \label{fig:SeqGAN}
\end{figure}

Yu \textit{et al.} proposed a sequential data generation framework \cite{Yu_seqgan_2017} that could address the issues with generating discrete data as previously mentioned in \ref{sec:time-series-gans}. This approach outperformed previous methods for generative modelling on real-world tasks, including;  a maximum likelihood estimation (MLE) trained LSTM, scheduled sampling \cite{bengio2015scheduled}, and Policy Gradient with bilingual evaluation understudy (PG-BLEU) \cite{papineni-etal-2002-bleu}. SeqGAN's generative model comprises RNNs with LSTM cells, and its discriminative model is a convolutional neural network (CNN). Given a dataset of structured sequences the authors train $G$ to produce a synthetic sequence $Y_{1:T} = (y_{1}...,y_{t}...,y_{T}), y_{t} \in \mathcal{Y}$ where $\mathcal{Y}$ is defined as the vocabulary of candidate tokens. $G$ is updated by a policy gradient and Monte Carlo (MC) search on the expected reward from $D$, see Figure \ref{fig:SeqGAN}. The authors used two datasets for their experiments. A Chinese poem dataset \cite{Chinese_Poems} and a Barack Obama Speech dataset \cite{Obama_RNN} with Adam optimisers and a batch size of 64. Their experiments are available online\footnote{SeqGAN GitHub: https://github.com/LantaoYu/SeqGAN/}.

Although the purpose of SeqGAN is to generate discrete sequential data, it opened the door to other GANs in generating continuous sequential and time series data. The authors use a synthetic dataset whose distribution is generated from a randomly initialised LSTM following a normal distribution. They also compare the generated data to real-world examples of poems, speech-language and music. SeqGAN showed competitive performance in generating the sequences and contributed heavily towards the further development of the continuous sequential GANs.

\subsubsection{Quant GAN (Jul. 2019)}
\begin{figure}[ht]
    \centering
    \includegraphics[width=0.5\columnwidth]{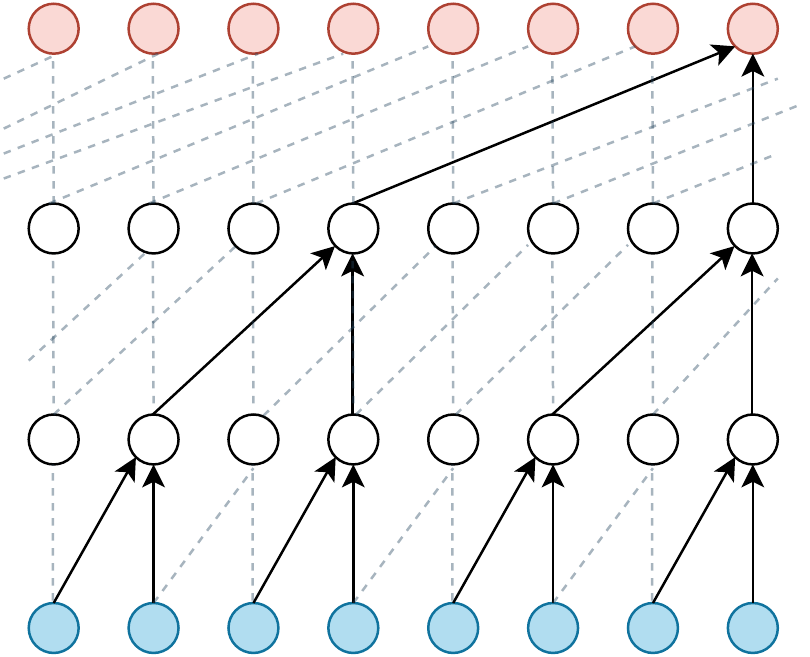}
    \caption{Dilated causal convolutional layer.}
    \label{fig:TCN_model}
\end{figure}

Quant GAN is a data-driven model that aims to capture long-range dependencies in financial time series data such as volatility clusters. Both the generator and discriminator use Temporal Convolutional Networks (TCN) with skip connections \cite{wiese_quantgan_2020} which are essentially dilated causal convolutional networks. They have the advantage of being suited to model long-range dependencies in continuous sequential data. The generator function is a novel stochastic volatility NN (SVNN) that consists of a volatility and drift TCN. Temporal blocks are the modules used in the TCN that consist of two dilated causal convolutions layers (Figure \ref{fig:TCN_model}) and two Parametric Rectified Linear Units (PReLU) as activation functions. Data generated by $G$ is passed to $D$ to produce outputs, which can then be averaged to give an MC estimate of $D$'s loss function. The authors used a dataset of daily spot-prices of the S\&P 500 from May 2009 until December 2018.

The authors aim to capture long-range dependencies in financial time series; however, modelling the series in continuous time over these long time frames would blow up the models' computational complexity. Therefore, this method models the time series in discrete time. The authors report that this approach is capable of outperforming more conventional models from mathematical finance (Constrained SVNN and generalised autoregressive conditional heteroskedasticity (GARCH) \cite{GARCH86}) but state that there remain issues that need to be resolved for this approach to become widely adopted. One such issue concerns the need for a unified metric for quantifying the performance of these GANs, a point we discuss further in Section \ref{sec:eval}.


\subsection{Continuous-variant GANs}

\subsubsection{Continuous RNN-GAN (C-RNN-GAN) (Nov. 2016)}
\label{sec:c-rnn-gan}
\begin{figure}[ht]
    \centering
    \includegraphics[width=0.85\columnwidth]{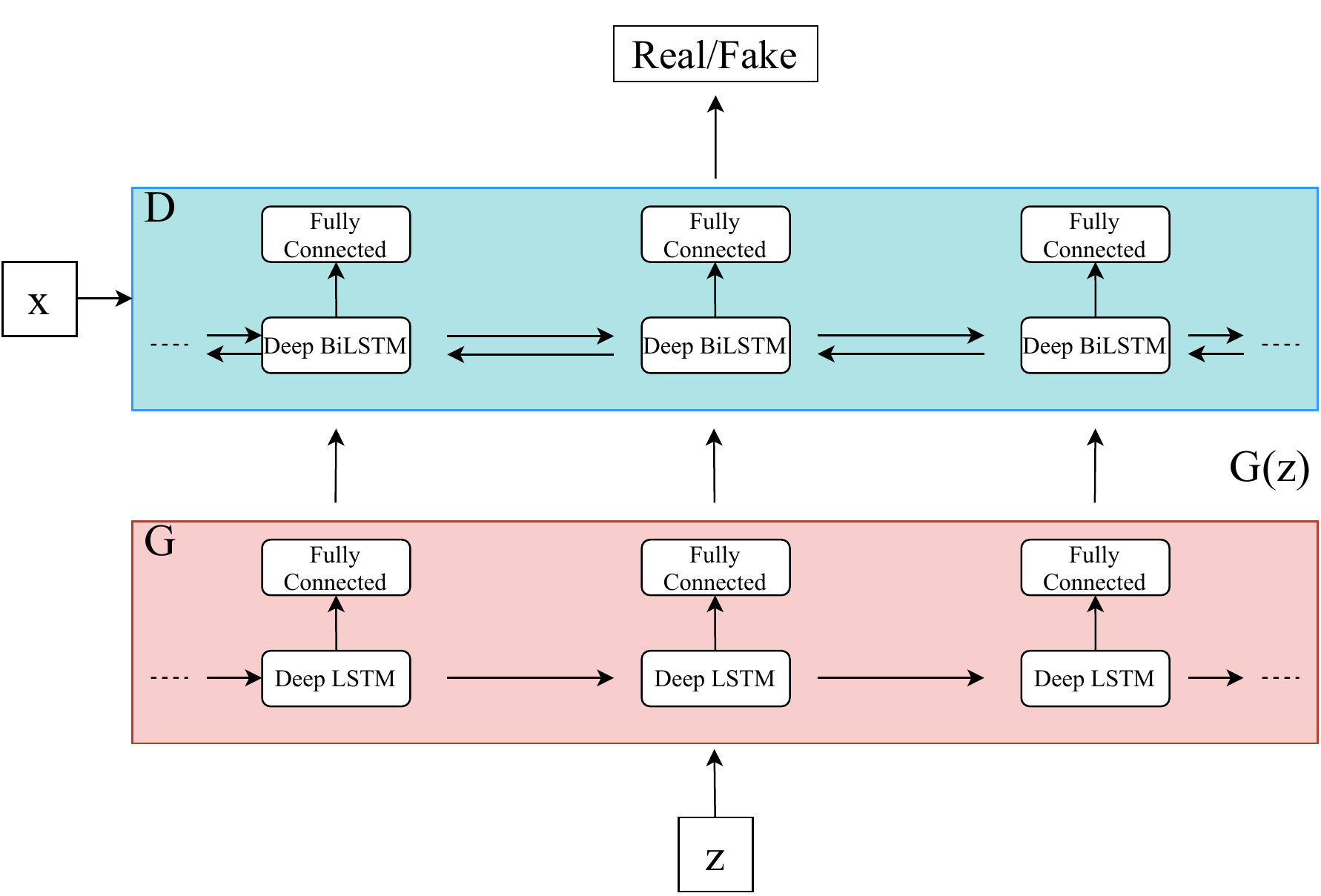}
    \caption{Structure of C-RNN-GAN's generator and discriminator.}
    \label{fig:C-RNN-GAN_model}
\end{figure}

In previous works, RNNs have been applied to modelling music but have generally used a symbolic representation to model this type of sequential data. Mogren proposed the C-RNN-GAN (Figure \ref{fig:C-RNN-GAN_model}), one of the first examples of using GANs to generate continuous sequential data\cite{mogren_c-rnn-gan_2016}. The generator is an RNN, and the discriminator a bidirectional RNN, which allows the discriminator to take the sequence context in both directions. The RNNs used in this work were two stacked LSTMs layers, with each cell containing 350 hidden units. The loss functions can be seen in (\ref{eq:C-RNN-GAN_G},\ref{eq:C-RNN-GAN_D}), where $z^{(i)}$ is a sequence of uniform random vectors in [0,1]$^k$, and $x^{(i)}$ is a sequence from the training data. $k$ is the dimensionality of the data in the random sequence.

\begin{equation}
{L}_{G} = \frac{1}{m} \sum_{i=1}^{m} log(1-D(G(z^{(i)})))
\label{eq:C-RNN-GAN_G}
\end{equation}

\begin{equation}
{L}_{D} = \frac{1}{m} \sum_{i=1}^{m} [-logD(x^{(i)}) - log(1-D(G(z^{(i)})))]
\label{eq:C-RNN-GAN_D}
\end{equation}

The C-RNN-GAN is trained with backpropagation through time (BPTT) and mini-batch stochastic gradient descent with L2 regularisation on the weights of both $G$ and $D$. Freezing was applied to both $G$ and $D$ when one network becomes too strong relative to the other. The dataset used was 3697 midi files from 160 different composers of classical music with a batch size of 20. Adam and Gradient Descent Optimisers were used during training; full implementation details are available online\footnote{C-RNN-GAN GitHub: https://github.com/olofmogren/c-rnn-gan/}. Overall the C-RNN-GAN was capable of learning the characteristics of continuous sequential data and, in turn, generate music. However, the author stated that their approach still needs work, particularly in rigorous evaluation of the generated data quality.


\subsubsection{Recurrent Conditional GAN (RCGAN) (2017)}
\begin{figure}[ht]
    \centering
    \includegraphics[width=\columnwidth]{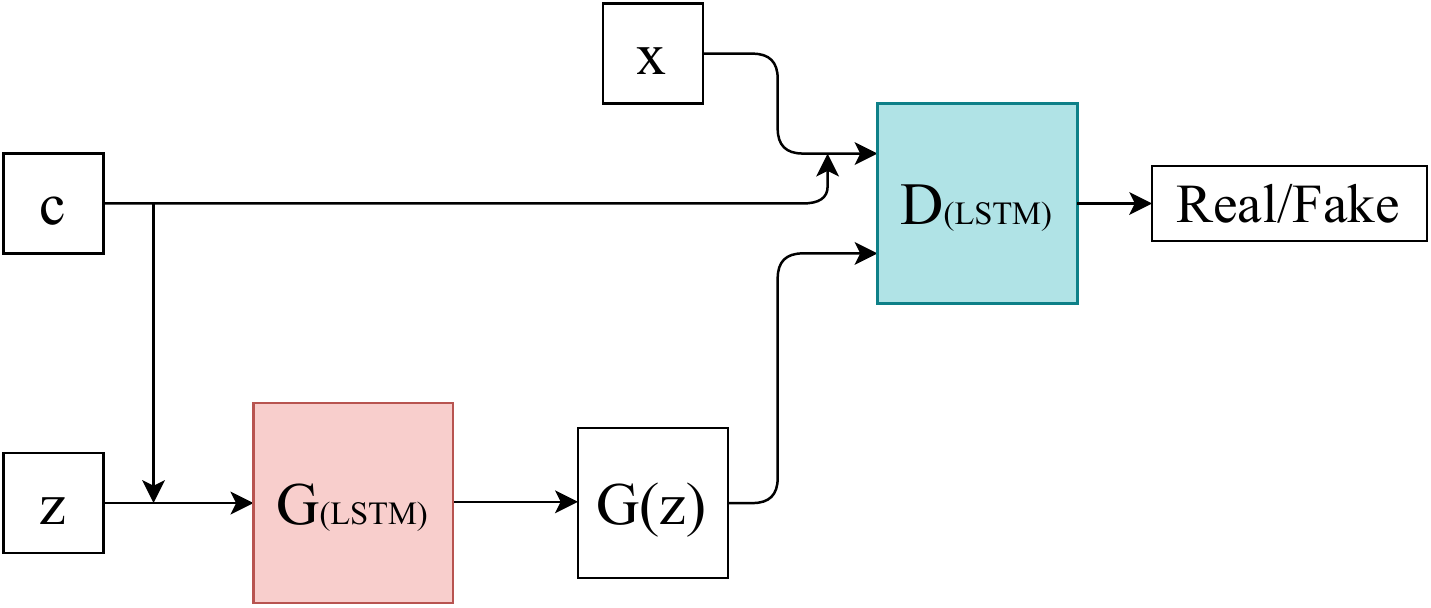}
    \caption{RCGAN architecture with conditional input \textbf{c}, input data \textbf{x} and latent variable \textbf{z}.}
    \label{fig:RCGAN_model}
\end{figure}

RCGAN for continuous data generation \cite{Esteban2017} differs architecturally from the C-RNN-GAN. Although the RNN LSTM is used, the discriminator is unidirectional, and the outputs of $G$ are not fed back as inputs at the next time step. There is also additional information that the model is conditioned on, which makes for a conditional RGAN; see the layout of the model in Figure \ref{fig:RCGAN_model}. The purpose of the RCGAN and RGAN in this work is to generate continuous time series with a focus on medical data intended for use in downstream tasks, and this was one of the first works in this area. The loss functions can be seen in Equations (\ref{eq:RCGAN_Dloss}, \ref{eq:RCGAN_Gloss}) where CE is the average cross-entropy between two sequences. $X_n$ are samples drawn from the training dataset. $y_n$ is the adversarial ground truth; for real sequences, it is a vector of 1s, and conversely, for generated or synthetic sequences, it is a vector of 0s. $Z_n$ is a sequence of points sampled from the latent space, and the valid adversarial ground truth is written here as \textbf{1}.

\begin{equation}
{D}_{loss}(X_n, y_n) = -CE(D(X_n), y_n)
\label{eq:RCGAN_Dloss}
\end{equation}

\begin{equation}
{G}_{loss}(Z_n) = D_{loss}(G(Z_n),\textbf{1}) =  -CE(D(G(Z_n)), \textbf{1})
\label{eq:RCGAN_Gloss}
\end{equation}

In the conditional case, the inputs to $D$ and $G$ are concatenated with some conditional information $c_n$. This variant of an RNN-GAN facilitates the generation of a synthetic continuous time series dataset with associated labels. Experiments were carried out on generated sine waves, smooth functions sampled from a Gaussian process with a zero-valued mean function, MNIST dataset as a sequence, and the Philips eICU database \cite{Phillips_eICU}. A batch size of 28 with Adam and Gradient Descent Optimisers were used for training. The authors propose a novel method for evaluating their model, which is discussed further in Section \ref{sec:eval}. Full experimental details can be found online\footnote{RCGAN GitHub: https://github.com/ratschlab/RGAN/}.


\subsubsection{Sequentially Coupled GAN (SC-GAN) (Apr. 2019)}
\begin{figure}[ht]
    \centering
    \includegraphics[width=0.85\columnwidth]{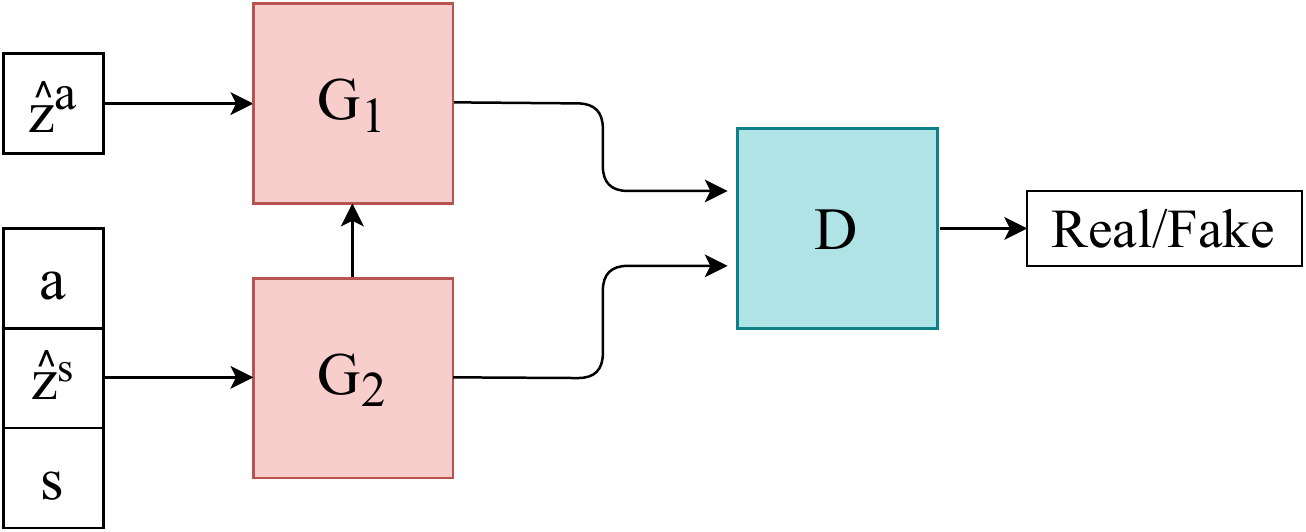}
    \caption{SC-GAN architecture.}
    \label{fig:SC-GAN_model}
\end{figure}

SC-GAN aims to generate patient-centric medical data to inform of a patient's current state and generate a recommended medication dosage based on the state \cite{wang_continuous_2019}. It consists of two coupled generators tasked with producing two outcomes, one for the current state of an individual and the other for a recommended medication dosage based on the individual's state. The discriminator is a two-layer bidirectional LSTM, and the coupled generators are both two-layer unidirectional LSTMs. See Figure \ref{fig:SC-GAN_model} for further details of the architecture.

$G_1$ generates the recommended medication dosage data ($\textbf{a}_1,\textbf{a}_2,...,\textbf{a}_T$) with a combined input of the sequential continuous patient state data ($\textbf{s}_0,\textbf{s}_1,...,\textbf{s}_{T-1}$) and a random noise sequence ($\hat{\textbf{z}}_{0}^{a},\hat{\textbf{z}}_{1}^{a},...,\hat{\textbf{z}}_{T-1}^{a}$) sampled from a uniform distribution. At each time step $t$ the input $\textbf{z}_{t}^{a}$ of $G_1$ is the concatenation of $\textbf{s}_t$ and $\hat{\textbf{z}}_{t}^{a}$. 

Conversely, $G_2$ is tasked with generating the patient state data $\textbf{s_t}$ and at each time step the input $\textbf{z}_{t}^{s}$ is the concatenation of $\textbf{s}_{t-1}$, $\textbf{a}_{t-1}$ and $\hat{\textbf{z}}_{t}^{s}$. This means that the outputs from $G_1$ and $G_2$ are the inputs to one another. Combining the generators together leads to the following loss function:

\begin{equation}
L_G = \frac{1}{N} \frac{1}{T} \sum_{i=1}^{N} \sum_{t=1}^{T} log(1-D(G(\textbf{z}_{i,t})))
\label{eq:SC-GAN_G}
\end{equation}

\begin{equation}
G(\textbf{z}_{i,t}) = [G_1(\textbf{z}_{i,t}^{a});G_2(\textbf{z}_{i,t}^{s})]
\label{eq:SC-GAN_Gz}
\end{equation}

Where N is the number of patients and T is the time length of the patient record. The SC-GAN has a supervised pretraining step for the generators to avoid an excessively strong $D$ that uses the least-squares loss. 

The discriminator is tasked with classifying the sequential patient-centric records as real or synthetic, and the loss function is defined as:

\begin{equation}
L_D = - \frac{1}{N} \frac{1}{T} \sum_{i=1}^{N} \sum_{t=1}^{T} (logD(\textbf{x}_{i,t}) + log(1-D(G(\textbf{z}_{i,t}))))
\label{eq:SC-GAN_D}
\end{equation}

where $\textbf{x}_{i,t} = [\textbf{s}_{t};\textbf{a}_{t}]$. This model contains novel coupled generators that coordinate to generate patient state and medication dosage data. It has performance close to real data for the treatment recommendation task. The dataset used in this experiment is MIMIC-III \cite{mimiciii}. The authors benchmark their SC-GAN against variants of  SeqGAN, C-RNN-GAN, and RCGAN and observe their model to be the best performing for this specific use case.


\subsection{Noise Reduction GAN (NR-GAN) (Oct. 2019)}
\label{sec:NR-GAN}
NR-GAN is designed for noise reduction in continuous time series signals but more specifically has been implemented for noise reduction in mice electroencephalogram (EEG) signals \cite{sumiya_nr-gan_2019}. This dataset was provided by the International Institute for Integrative Sleep Medicine (IIIS). EEG is the measure of the brain's electrical activity, and it commonly contains significant noise artefact. NR-GAN's core idea is to reduce or remove the noise present in the frequency domain representation of an EEG signal. The architecture of $G$ is a two-layer 1-D CNN with a fully connected layer at the output. $D$ contains almost the same two-layer 1-D CNN structure with the fully-connected layer replaced by a softmax layer to calculate the probability the input belongs to the training set. The loss functions are defined in (\ref{eq:NRGAN-G},\ref{eq:NRGAN-D}) as:

\begin{equation}
{G}_{loss}= \sum_{x\in S_{ns}} [log(1-D(G(x))) + \alpha \|x - G(x)\|^2]
\label{eq:NRGAN-G}
\end{equation}

\begin{equation}
{D}_{loss}= \sum_{x\in S_{ns}} [log(D(G(x)))] + \sum_{y\in S_{cs}}[log(1-D(y))]
\label{eq:NRGAN-D}
\end{equation}

where $S_{ns}$ and $S_{cs}$ are the noisy and clear EEG signals, respectively. $\alpha$ is a hyperparameter that essentially controls the aggressiveness of noise reduction; the authors chose a value of $\alpha = 0.0001$. 

\begin{figure}[ht]
    \centering
    \includegraphics[width=\columnwidth]{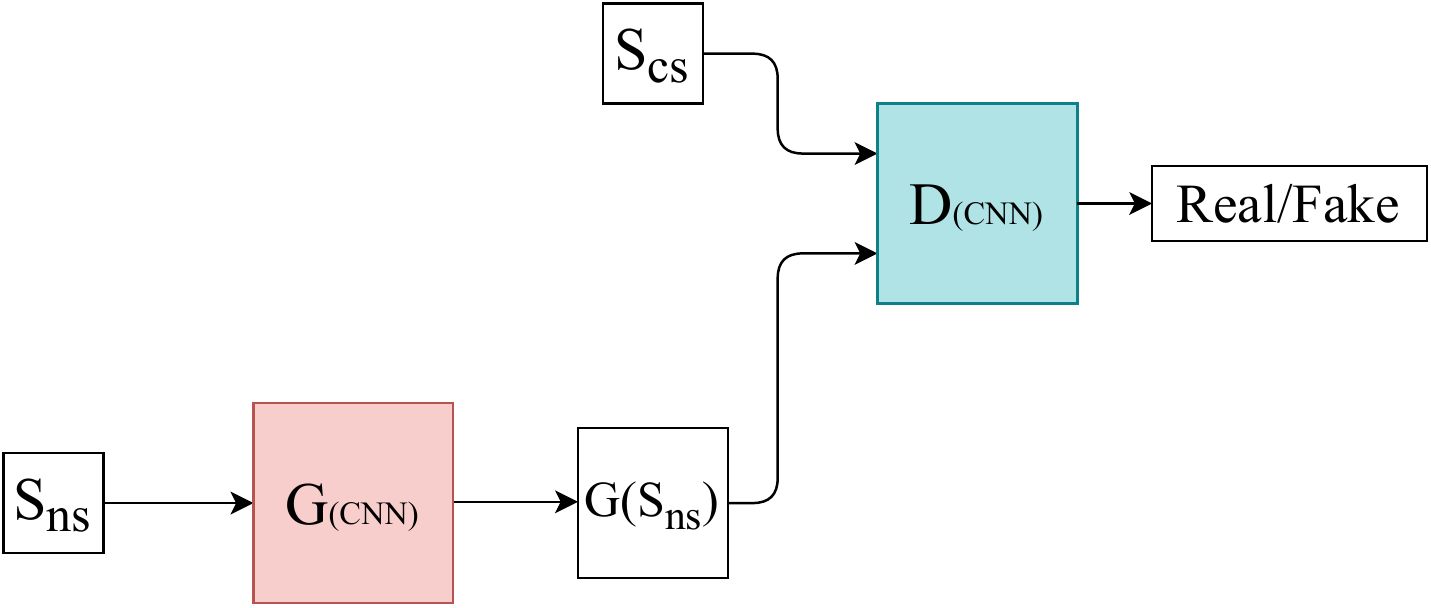}
    \caption{NR-GAN architecture with noisy EEG input ${S_{ns}}$, clean input data ${S_{cs}}$.}
    \label{fig:NRGAN_model}
\end{figure}

For this work, the generator does not sample from a latent space; rather, it attempts to generate the clear signal from the noisy EEG signal input, see Figure \ref{fig:NRGAN_model}. The authors found that the NR-GAN is competitive with classical frequency filters in terms of noise reduction. They also state that the experimental conditions may favour the NR-GAN and list some limitations in terms of the amount of noise NR-GAN can handle and the influence of $\alpha$. However, this is a novel method for noise reduction in continuous sequential data using GANs.

\subsubsection{Time GAN (Dec. 2019)}
\begin{figure}[ht]
    \centering
    \includegraphics[width=\columnwidth]{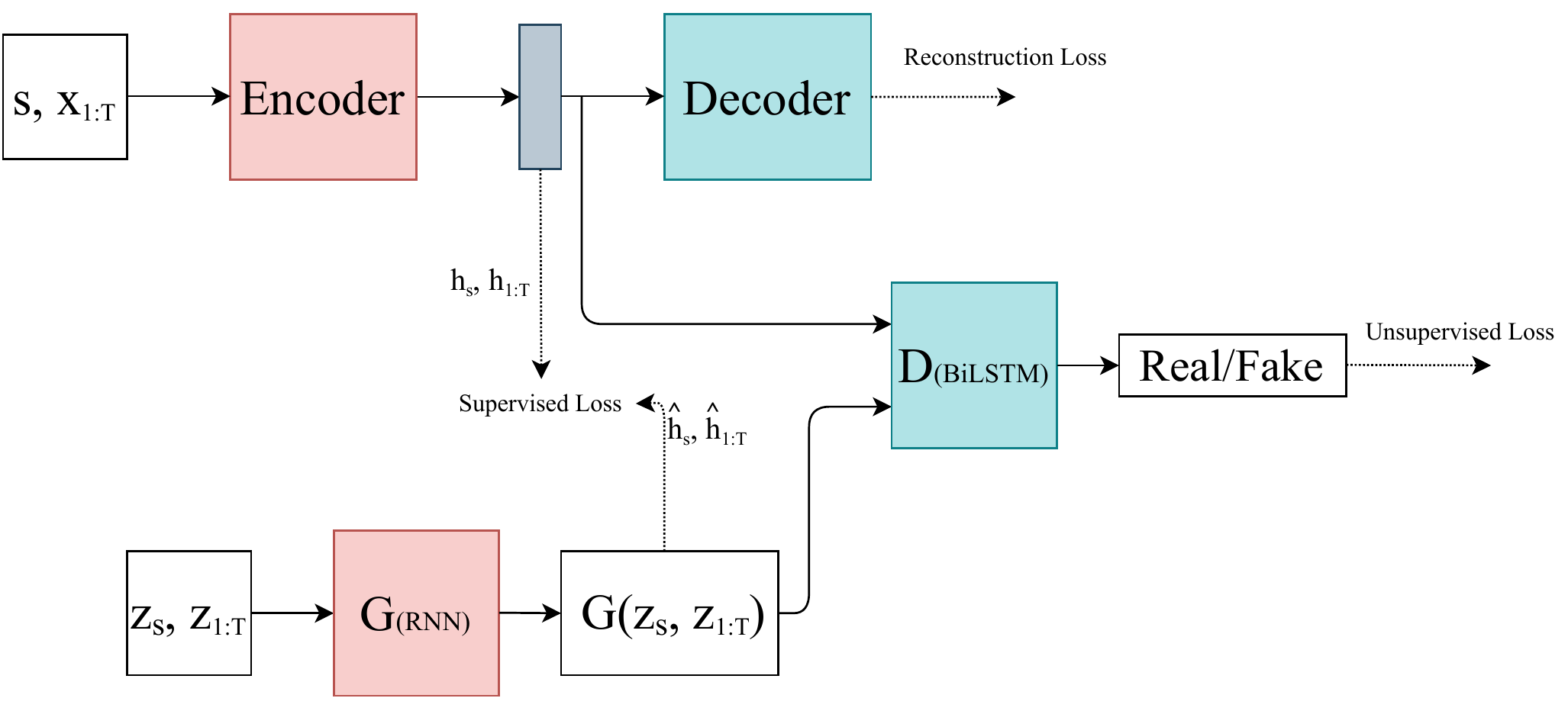}
    \caption{TimeGAN architecture.  }
    \label{fig:TimeGAN_model}
\end{figure}

TimeGAN provides a framework that utilises both the conventional unsupervised GAN training method and the more controllable supervised learning approach \cite{yoon_timeseriesGAN_2019}. By combining an unsupervised GAN network with a supervised autoregressive model, the network aims to generate time series with preserved temporal dynamics. The architecture of the TimeGAN framework is illustrated in Figure \ref{fig:TimeGAN_model}. The input to the framework is considered to consist of two elements, a static feature and a temporal feature. \textbf{s} represents a vector of static features and \textbf{x} of temporal features at the input to the encoder. The generator takes a tuple of static and temporal random feature vectors drawn from a known distribution. The real and synthetic latent codes $\textbf{h}$ and $\hat{\textbf{h}}$  are used to calculate the supervised loss element of this network. The discriminator receives the tuple of real and synthetic latent codes and classifies them as either real ($y$) or synthetic ($\hat{y}$), the $\Tilde{}$ operator denotes the sample as either real or fake.

The three losses used in TimeGAN are calculated as follows:

\begin{equation}
{L}_{reconstruction} = \mathbb{E}_{s,x_{1:T} \sim p}[ \|s-\Tilde{s}\|_{2} + \sum_{t} \| x_{t} - \Tilde{x}_{t}\|_{2}] 
\label{eq:TimeGAN_R-loss}
\end{equation}

\begin{equation}
{L}_{unsupervised} = \mathbb{E}_{s,x_{1:T} \sim p}[log(y_{S}) + \sum_{t} log(y_{t}) ] + \mathbb{E}_{s,x_{1:T} \sim{\hat{p}}} [log(1- \hat{y}_{S}) + \sum_{t} log(1- \hat{y}_{t})]
\label{eq:TimeGAN_U-loss}
\end{equation}

\begin{equation}
{L}_{supervised} = \mathbb{E}_{s,x_{1:T} \sim p}[ \sum_{t} \|h_{t} -g_{X}(h_{S}, h_{t-1}, z_{t})\|_{2}]
\label{eq:TimeGAN_S-loss}
\end{equation}

The creators of TimeGAN conducted experiments on generating sine waves, stocks (daily historical Google stocks data from 2004 to 2019), energy (UCI Appliances energy prediction dataset) \cite{UCIArchive2017}, and event (private lung cancer pathways dataset) datasets. A batch size of 128 and Adam optimiser were used for training, implementation details are available online\footnote{TimeGAN GitHub: https://github.com/jsyoon0823/TimeGAN}. The authors demonstrated improvements over other state-of-the-art time series GANs such as  RCGAN C-RNN-GAN and WaveGAN.


\subsubsection{Conditional Sig-Wasserstein GAN (SigCWGAN) (Jun. 2020) }
A problem addressed by \cite{ni_sig-WGAN_2020} is that long time series data streams can greatly increase the dimensionality requirements of generative modelling, which may render such approaches infeasible. To counter this problem, the authors develop a metric named Signature Wasserstein-1 (Sig-$W_1$) that captures time series models' temporal dependency and uses it as a discriminator in a time series GAN. It provides an abstract and universal description of complex data streams and does not require costly computation like the Wasserstein metric. A novel generator is also presented that is named conditional autoregressive feed-forward neural network (AR-FNN) that captures the auto-regressive nature of the time series. The generator is capable of mapping past series and noise into future series. For a rigorous mathematical description of their method, the interested reader should consult \cite{ni_sig-WGAN_2020}.

For the AR-FNN generator the idea is to obtain the step-q estimator $\hat{X}^{(t)}_{t+1:t+q}$. The loss function for $D$ is defined as:

\begin{equation}
L(\theta)= \sum_{t}|\mathbb{E}_{\mu}[S_M(X_{t+1:t+q})|X_{t-p+1:t}] - \mathbb{E}_v[S_M(\hat{X}_{t+1:t+q}^{(t)})|X_{t-p+1:t}]|
\label{eq:SIGCWGAN_D}
\end{equation}

Where $v$ and $\mu$ are the conditional distributions induced by the real data and synthetic generator, respectively, further details of the author's algorithm can be found in the appendix of the original paper.
The authors state that SigCWGAN eliminates the problem of approximating a costly $D$ and simplifies training. It is reported to achieve state-of-the-art results on synthetic and empirical datasets compared to TimeGAN, RCGAN and Generative Moment Matching Networks (GMMN) \cite{li2015GMMN}. The empirical dataset consists of the S\&P 500 index (SPX) and Dow Jones index (DJI) and their realized volatility, which is retrieved from the Oxford-Man Institute’s "realised library" \cite{Oxford_Man}. A batch size of 200 with the Adam optimiser was used for training\footnote{SigCWGAN GitHub: https://github.com/SigCGANs/Conditional-Sig-Wasserstein-GANs/}.


\subsubsection{Decision Aware Time series conditional GAN (DAT-CGAN) (Sept. 2020)}
This framework is designed to provide support for end-users decision processes, specifically in financial portfolio choices. It uses a multi-Wasserstein loss on structured decision-related quantities \cite{sun_decision-aware_2020}. The discriminator loss and generator loss are defined in Equations (\ref{eq:DAT-CGAN_D}) and (\ref{eq:DAT-CGAN_G}) respectively. For further details on the loss functions, see Section 3 of the original paper.

\begin{equation}
\mathop{inf}_{\eta} \mathop{sup}_{\gamma_{k},\theta_{j,k}} \sum_{k=1}^{K} \omega_k(\mathbb{E}_{k}^r - \mathbb{E}_{k}^{G_{\eta}})+ \sum_{k=1}^{K} \sum_{j=1}^{J} \lambda_{j,k} (\mathbb{E}_{j,k}^{f,R} - \mathbb{E}_{j,k}^{f, G_{\eta}}) 
\label{eq:DAT-CGAN_D}
\end{equation}

\begin{equation}
\mathop{inf}_{\eta} - \sum_{k} \omega_k\mathbb{E}_{k}^{G_{\eta}} - \sum_{k,j} \lambda_{j,k}  \mathbb{E}_{j,k}^{f, G_{\eta}}
\label{eq:DAT-CGAN_G}
\end{equation}

The generator is a two-layer feed-forward neural network for each input which are assets in this case. $G$ outputs asset returns that are used to compute decision-related quantities. These quantities are fed into $D$, which is also a two-layer feed-forward NN. Further details about the architecture can be found in the appendix of \cite{sun_decision-aware_2020}. The dataset used is daily price data for each of four U.S. Exchange-traded fund (ETFs), i.e. Material (XLB), Energy (XLE), Financial (XLF) and Industrial (XLI) ETFs, from 1999 to 2016. The authors found this model capable of high-fidelity time series generation that supports decision processes by end-users due to incorporating a decision-aware loss function. However, this approach's limitation is that the computational complexity of this model is vast and requires one month of training time for a single generative model.






\subsection{Synthetic biomedical Signals GAN (SynSigGAN) (Dec. 2020)}
\begin{figure}[ht]
    \centering
    \includegraphics[width=0.8\columnwidth]{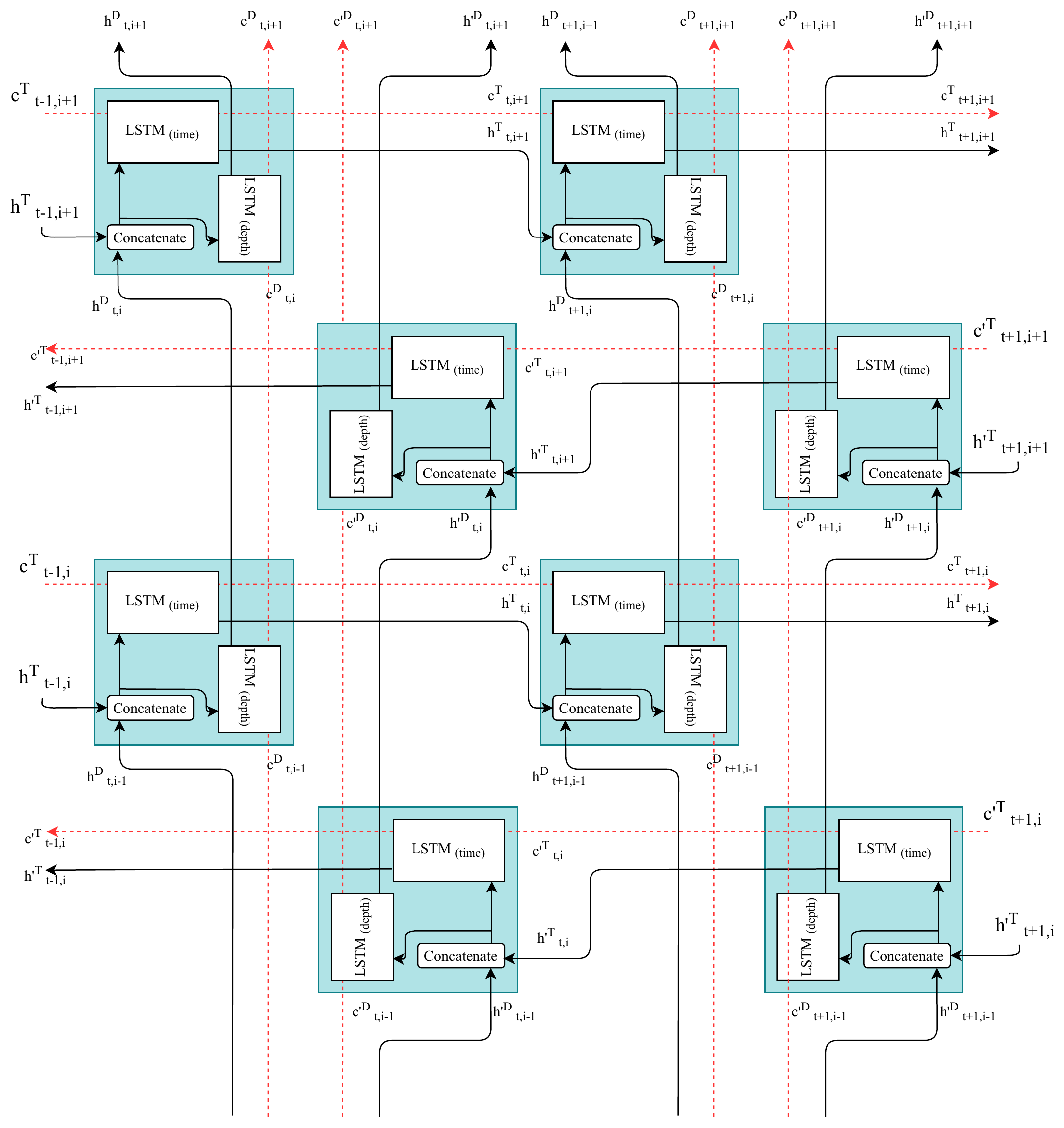}
    \caption{Architecture of BiGridLSTM with LSTM blocks for the time and depth dimension, $`$ symbol indicates reverse in the figure as in \cite{Fei_BiGridLSTM_2018}.}
    \label{fig:BiGridLSTM}
\end{figure}

SynSigGAN is designed to generate different kinds of continuous physiological/biomedical signal data \cite{hazra_synsiggan_2020}. It is capable of generating electrocardiogram (ECG), electroencephalogram (EEG), electromyography (EMG), and photoplethysmography (PPG) from MIT-BIH Arrhythmia database \cite{MITBIH_Arrhythmia}, Siena Scalp EEG database \cite{Sienna_Scalp_EEG} and BIDMC PPG and Respiration dataset \cite{BIDMC_PPG}. A novel GAN architecture is proposed here that uses a bidirectional grid long short term memory (BiGridLSTM) for the generator (Figure \ref{fig:BiGridLSTM}) and a CNN for the discriminator. The BiGridLSTM is a combination of a double GridLSTM (a version of LSTM that can represent the LSTMs in a multidimensional grid) with two directions that can combat the gradient phenomenon from two dimensions and found to work well in time sequence problems. The authors used the value function defined previously in Equation (\ref{eq:1}).

SynSigGAN is capable of capturing the different physiological characteristics associated with each of these signal types and has demonstrated an ability to generate biomedical time series data with a max sequence length of 191 data points. The authors also present a preprocessing stage to clean and refine the biomedical signals in this paper. They compare their architecture to several variants (BiLSTM-GRU, BiLSTM-CNN GAN, RNN-AE GAN, Bi-RNN, LSTM-AE, BiLSTM-MLP, LSTM-VAE GAN, and RNN-VAE GAN) and found the BiGrid-LSTM as the best performing model.  


\section{Applications}
\label{sec:applications}
\subsection{Data Augmentation}
It is common knowledge in the deep learning community that GANs are among the methods of choice when discussing data augmentation. Reasons for augmenting datasets range from increasing the size/variety of small and imbalanced datasets \cite{nikolaidis2019augmenting, abdelfattah_augmenting_2018, haradal_biosignal_2018, kiyasseh_plethaugment_2020} to reproducing restricted datasets for dissemination.

A well-defined solution to the data shortage problem is transfer learning, and it works well in domain adaptation which has led to advancements in classification and recognition problems \cite{Pan2010_transferlearning}. However, it has been found that augmenting datasets with GANs can lead to further improvements in certain classification and recognition tasks \cite{zhang2019snoregans}. Data synthesised by a GAN can adhere to stricter privacy measures discussed in Section \ref{sec:privacy}. This further demonstrates the advantages of augmenting your training dataset with GANs over implementing transfer learning with a pre-trained model from a different domain on a smaller dataset.

Many researchers find that accessing datasets for their deep learning research and models to be time-consuming, laborious work, particularly when the research is concerned with personal sensitive data. Often medical and clinical data are presented as continuous sequential data that can only be accessed by a small contingent of researchers who are not at liberty to disseminate their research openly. This, in turn, may lead to stagnation in the research progress in these domains. 

\begin{figure}[ht]
    \centering
    \includegraphics[width=0.49\columnwidth]{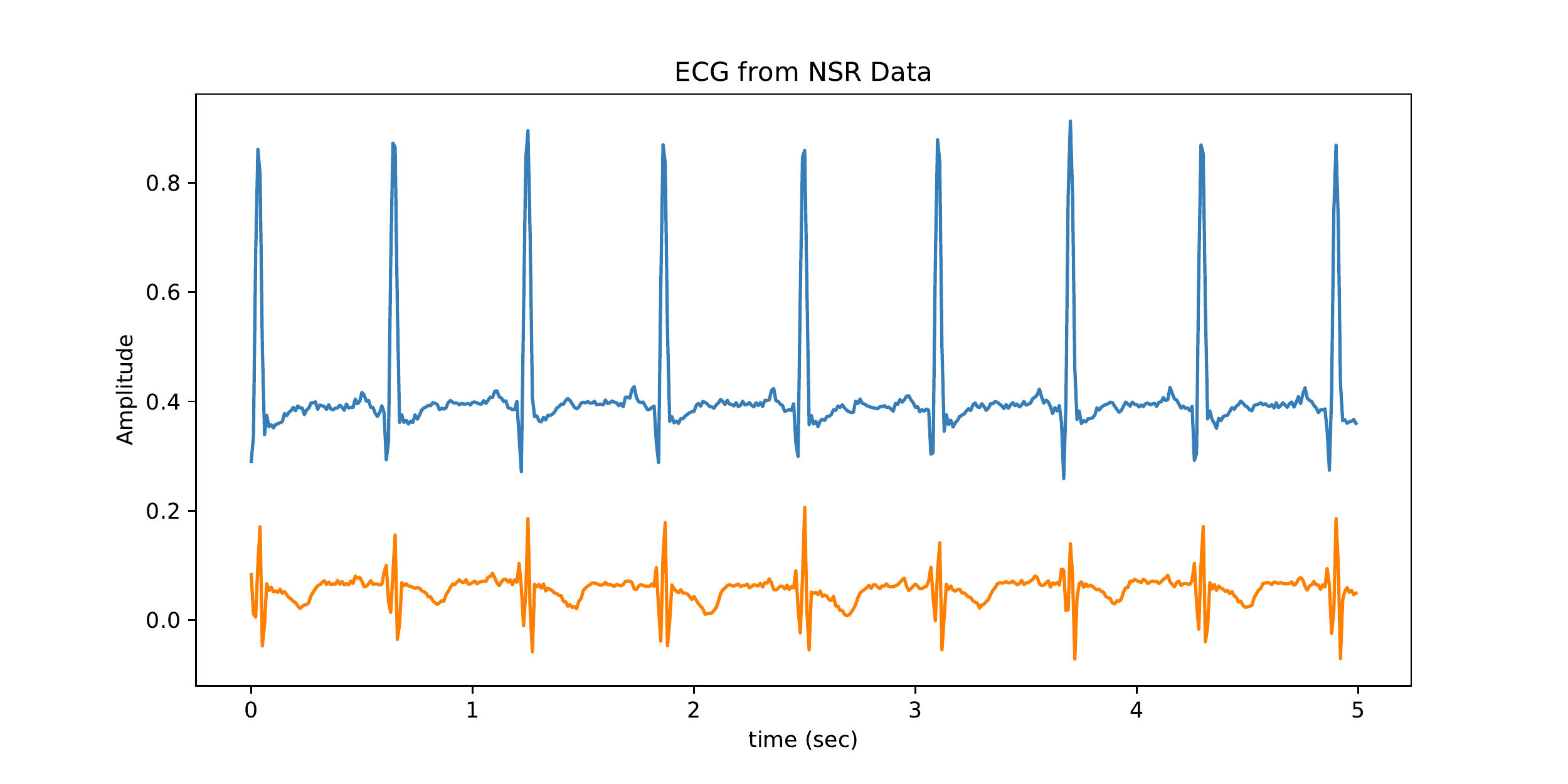}
    \includegraphics[width=0.49\columnwidth]{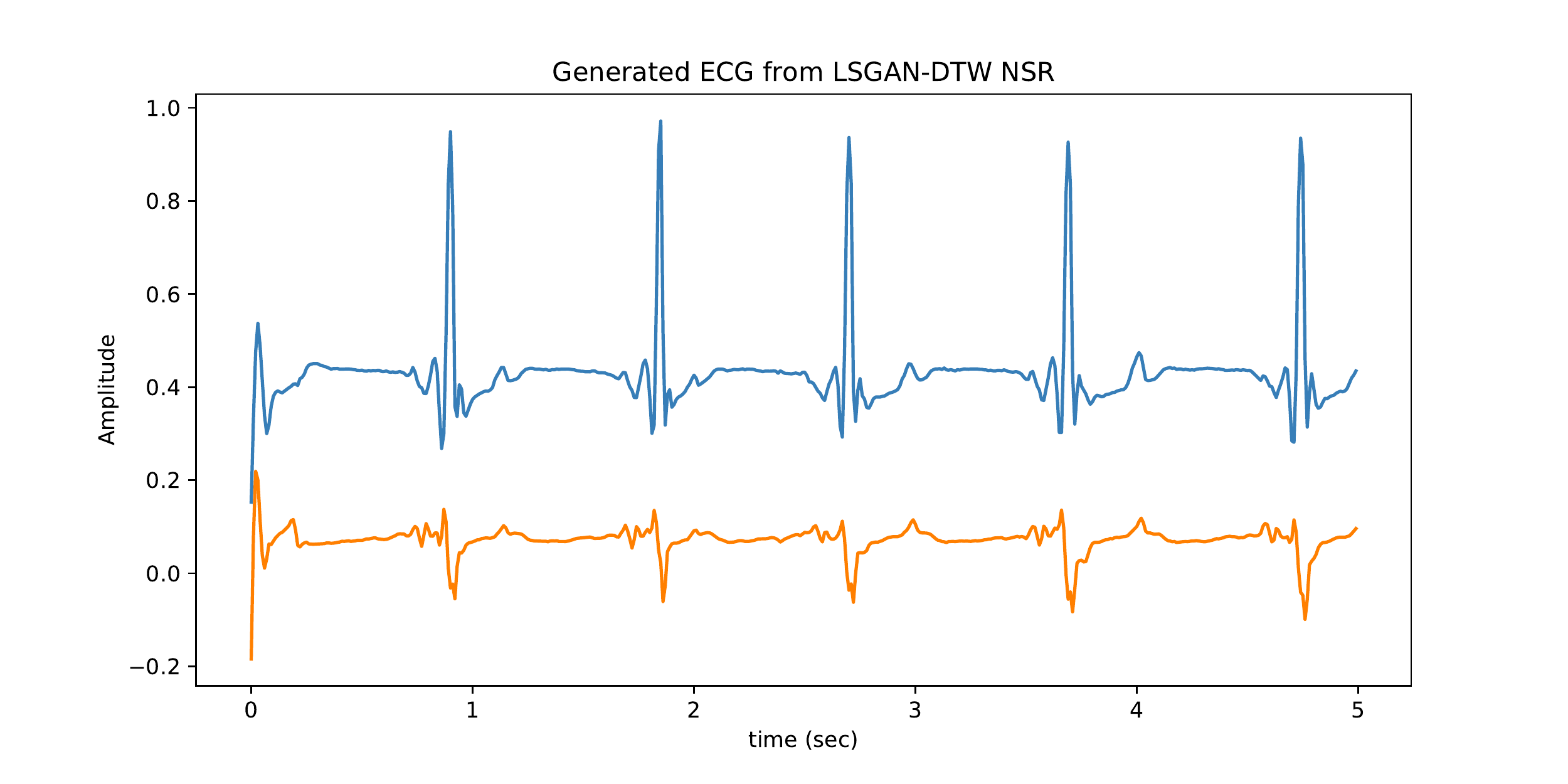}
    \caption{An example of dependent multichannel ECG data (left) and  generated ECG from a multivariate GAN (right) \cite{Brophy2020}. NSR indicates the training dataset which is the Normal Sinus Rhythm. The generated data is produced by a GAN named by the authors as LSGAN-DTW.}
    \label{fig:Generated_ECG}
\end{figure}

Fortunately, we are beginning to see the uptake of GANs applied to time series with these types of medical and physiological data  \cite{Esteban2017, hazra_synsiggan_2020, Delaney2019, Zhu_lstmcnn_2019, brophy_quick_2019}. With \cite{Brophy2020} showing dependent multivariate continuous high-fidelity physiological signal generation is capable via GANs, demonstrating the impressive capability of these networks. See Figure \ref{fig:Generated_ECG} for an example of the real input and synthetic generated data. 

Of course, this is not comprehensive coverage of the research using time series GANs for data synthesis and augmentation. GANs have been applied to time series data for a plethora of use cases. 

Audio generation (both music and speech) and text-to-speech (TTS) \cite{juvela_tts_2019} has been a popular area for researchers to explore with GANs. The C-RNN-GAN described in Section \ref{sec:c-rnn-gan} was one of the seminal works to apply GANs to generating continuous sequential data in the form of music.

In the financial sector, GANs have been implemented to generate data and predict/forecast values. Wiese \textit{et al.} implemented a GAN to approximate financial time series in discrete-time \cite{wiese_quantgan_2020}. In \cite{ni_sig-WGAN_2020}, the authors designed a decision-aware GAN that generates synthetic data and supports decision processes to financial portfolio selection of end-users. 

Other time series generation/prediction methods range from estimating soil temperature \cite{li_gans-lstm_2020} to predicting medicine expenditure based on the current state of patients \cite{kaushik_medicine_2020}.

\subsection{Imputation}
\begin{figure}[ht]
    \centering
    \includegraphics[width=0.5\columnwidth]{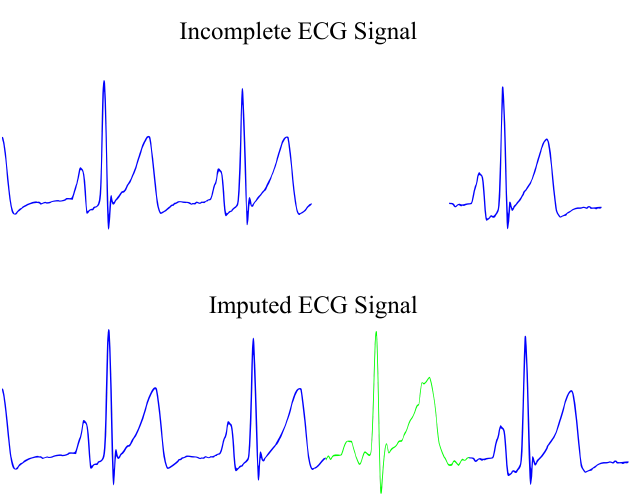}
    \caption{An example of the incomplete corrupted time series (top) and imputed signal (bottom).}
    \label{fig:MTS_imputation}
\end{figure}

In real-world datasets, missing or corrupt data is an all too common problem that leads to downstream problems. These issues manifest themselves in further analytics of the dataset and can induce biases in the data. Common methods of dealing with missing or corrupted data in the past have been the deletion of data streams containing the missing segments, statistical modelling of the data, or machine learning imputation approaches. Looking at the latter, we review the work in imputing these data using GANs. Guo {et al.} designed a GAN based approach for multivariate time series imputation \cite{guo_datamts_2019}, see Figure \ref{fig:MTS_imputation} for an example of imputed data from a toy experiment \cite{brophy_quick_2019}.

\subsection{Denoising}
Artefacts induced in time series data often manifest themselves as noise in the signals. This has become an ever-present challenge in further processing and analytical applications. Corrupted data can cause biases in the datasets or lead to degradation in the performance of critical systems such as those used for health applications. Common methods for dealing with noise include the use of adaptive linear filtering. Another approach recently explored in \cite{sumiya_nr-gan_2019} used GANs as a noise-reduction technique in EEG data. Their experiments showed that their proposed NR-GAN (Section \ref{sec:NR-GAN}) was capable of competitive noise reduction performance compared to more traditional frequency filters.  


\subsection{Anomaly Detection}
Detecting outliers or anomalies in time series data is an important part of many real-world systems and sectors. Whether it is detecting unusual patterns in physiological data that may be a precursor to some more malicious condition or detecting irregular trading patterns on the stock exchange, anomaly detecting can be vital to keeping us informed on important information. Statistical measures of non-stationary time series signals may achieve good performance on the surface, but they might also miss some important outliers present in deeper features. They may also struggle in exploiting large unlabelled datasets; this is where the unsupervised deep learning approaches can outperform the conventional methods. Zhu \textit{et al.} designed a GAN algorithm for anomaly detection in time series data (ECG and taxi dataset) with LSTMs and GANs, which achieved superior performance compared to conventional, more shallow approaches. Similar approaches have been applied to detect cardiovascular diseases \cite{luer_anomalyGAN_2019}, in cyber-physical systems to detect nefarious players \cite{tetko_mad-gan_2019} and even irregular behaviours such as stock manipulation on the stock markets  \cite{leangarun_stock-anom_2018}.

\subsection{Other Applications}
Some works have utilised image-based GANs for time series and sequential data generation by first converting their sequences to images via some transformation function and training the GAN on these images. Once the GAN converges, similar images can be generated; then, a sequence can be retrieved using the inverse of the original transformation function. For example, this approach has been implemented in audio generation with waveforms \cite{donahue2019WaveGAN, kolokolova_gans_2020, cheng_vTGAN_2020}, anomaly detection \cite{choi_gan-based_2020} and physiological time series generation \cite{brophy_quick_2019}.   

\section{Evaluation Metrics}
\label{sec:eval}
As mentioned in Section \ref{sec:gan} GANs can be difficult to evaluate, and researchers are yet to agree on what metrics reflect the GANs performance best. There has been plenty of metrics proposed in the literature \cite{borji_pros_2018} with most of them suited to the computer vision domain. Work is still ongoing to suitably evaluate time series GANs. We can break down evaluation metrics into two categories: qualitative and quantitative. Qualitative evaluation is another term for human visual assessment via the inspection of generated samples from the GAN. However, this cannot be deemed a full evaluation of GAN performance due to the lack of a suitable \textit{objective} evaluation metric. The quantitative evaluation includes the use of metrics associated with statistical measures used for time series analytics and similarity measures such as; Pearson Correlation Coefficient (PCC), percent root mean square difference (PRD), (Root) Mean Squared Error MSE and RMSE, Mean Relative Error (MRE) and Mean Absolute Error (MAE). These metrics are among the most commonly used for time series evaluation and, as such, used as a suitable GAN performance metric as they can reflect the stability between the training data and synthetic generated data. 

Several metrics have become well-established choices in evaluating image-based GANs, and some of these have permeated through to the sequential and time series GANs such as Inception Score (IS) \cite{salimans_IS_2016}, Fr\'{e}chet (Inception) Distance (FD and FID) \cite{heusel2018FID}. Structural Similarity Index (SSIM) is a measure of similarity between two images. However, \cite{parthasarathy_controlled_2020} use this with time series data as SSIM does not exclude itself from comparing aligned sequences of fixed length. Of course, some of these metrics are measures of similarities/dissimilarities between two probability distributions, suitable for many types of data, particularly the maximum mean discrepancy (MMD) \cite{gretton_twosample_2012} is very suitable to this task across domains. Another metric that generalises well to the sequential data case is the Sliced-Wasserstein Distance approximating the Wasserstein distance by computing Wasserstein distances between all 1d-projections of two distributions.

The data generated from GANs have been used in downstream classification tasks. Using the generated data together with the training data has lead to the Train on Synthetic, Test on Real (TSTR) and Train on Real, Test on Synthetic (TRTS) evaluation methods, first proposed by Esteban \textit{et al.} \cite{Esteban2017}. In scoring downstream classification applications that use both real and generated data, studies have adopted the precision, recall, and F1 scores to determine the classifier's quality and, in turn, the quality of the generated data. Other accuracy measures of classifier performance include the weighted accuracy (WA) and unweighted average recall (UAR).

Often used distance and similarity measures in time series data are the Euclidean Distance (ED) and Dynamic Time Warping (DTW) algorithms. Multivariate (in)dependent DTW (MVDTW), implemented in \cite{Brophy2020}, can determine similarity measures across both dependent and independent multichannel time series signals.

Other metrics used across different applications include: 

\begin{itemize}
  \item \textbf{Financial Sector}; autocorrelation function (ACF) score, DY metric. 
  \item \textbf{Temperature Estimation}; Nash-Sutcliffe model efficiency coefficient (NS), Willmott index of agreement (WI) and the Legates and McCabe index (LMI).
  \item \textbf{Audio Generation}; Normalised Source-to-Distortion Ratio (NSDR), Source-to Interference Ratio (SIR), Source-to-artifact ratio (SAR) and t-SNE \cite{tSNE_2008}.
\end{itemize}

For a full list of GAN architectures reviewed in this work, their applications, evaluation metrics, and datasets used in their respective experiments, see Table \ref{table:GANs}. Results for the sine wave and ECG generation using variants of GAN architectures can be found in Tables \ref{table:GAN-res-SINE} and \ref{table:GAN-res-MITBIH}, respectively.

\begin{table}[ht]
    \centering
    \caption{A list of GAN architectures, their applications, and datasets used in their experiments and evaluation metrics used to judge the quality of the respective GANs. For novel approaches, the GAN name is given as they have been covered already in Section \ref{sec:time-series-gans}} 
    \label{table:GANs}
    \begin{tabular}{  p{0.2\columnwidth}  p{0.25\columnwidth}  p{0.2\columnwidth} p{0.2\columnwidth} }
        \toprule
\textbf{Application}      
& \textbf{GAN Architecture(s)}   
& \textbf{Dataset(s)}
& \textbf{Evaluation Metrics} \\\midrule
Medical/Physiological Generation 
& LSTM-LSTM, \cite{Esteban2017} \cite{nikolaidis2019augmenting} \cite{abdelfattah_augmenting_2018}, \cite{haradal_biosignal_2018}, \cite{harada_biosignal_2019}, \cite{wang_continuous_2019} 

LSTM-CNN, \cite{Brophy2020} \cite{Delaney2019}

BiLSTM-CNN, \cite{Zhu_lstmcnn_2019}

BiGridLSTM-CNN, \cite{hazra_synsiggan_2020} 

CNN-CNN, \cite{Hartmann2018}, \cite{fahimi_towards_2019}

AE-CNN, \cite{pascual_epilepsygan_2020}

FCNN \cite{yi_adversarial_2019}

& EEG, ECG, EHRs, PPG, EMG, Speech, NAF, MNIST, Synthetic sets
& TSTR, MMD, Reconstruction error, DTW, PCC, IS, FID, ED, S-WD, RMSE, MAE, FD, PRD, Averaging Samples, WA, UAR, MV-DTW\\\hline

Financial time series generation/prediction
&TimeGAN \cite{yoon_timeseriesGAN_2019} 

SigCWGAN \cite{ni_sig-WGAN_2020}

DAT-GAN \cite{sun_decision-aware_2020}

QuantGAN \cite{wiese_quantgan_2020}
& S\&P500 index (SPX), Dow Jones Index (DJI), ETFs
& Marginal Distributions, Dependencies, TSTR, Wasserstein Distance, EM distance, DY Metric, ACF score, leverage effect score, discriminative score, predictive score \\\hline

Time series Estimation/Prediction      
& LSTM-NN \cite{li_gans-lstm_2020}

LSTM-CNN \cite{kaushik_medicine_2020} 

LSTM-MLP \cite{kaushik_medicine_2020}                       
& Meteorological data, Truven MarketScan dataset
& RMSE, MAE, NS, WI, LMI\\\hline

Audio Generation      
& C-RNN-GAN  \cite{mogren_c-rnn-gan_2016}

TGAN (variant) \cite{cheng_vTGAN_2020} 

RNN-FCN \cite{zhang_g-rnn-gan_2020} 

DCGAN (variant) \cite{kolokolova_gans_2020}

CNN-CNN \cite{juvela_tts_2019}
& Nottingham dataset, Midi music files, MIR-1K, TheSession, Speech
& Human perception, Polyphony, Scale Consistency, Tone Span, Repetitions, NSDR, SIR, SAR, FD, t-SNE, Distribution of notes\\\hline

Time series Imputation/Repairing   
& MTS-GAN  \cite{guo_datamts_2019}

CNN-CNN \cite{qu_novel_2020} 

DCGAN variant \cite{han_content-aware_2020}

AE-GRUI \cite{luo_e2gan_2019}

RGAN \cite{sun_exploration_2018}

FCN-FCN \cite{chen_traffic_2020}

GRUI-GRUI \cite{luo_multivariate_2018}

& TEP, Point Machine, Wind Turbin data, PeMS, PhysioNet Challenge 2012, KDD CUP 2018, Parking lot data, 
& Visually, MMD, MAE, MSE, RMSE, MRE, Spatial Similarity, AUC score\\\hline

Anomaly Detection      
& LSTM-LSTM  \cite{leangarun_stock-anom_2018}

LSTM-(LSTM\&CNN) \cite{zhu_lstmgan_anom_2019} 

LSTM-LSTM (MAD-GAN) \cite{tetko_mad-gan_2019} 

& SET50, NYC Taxi data, ECG, SWaT, WADI
& Manipulated data used as a test set, ROC Curve, Precision, Recall, F1, Accuracy\\\hline

Other time series generation 
& VAE-CNN  \cite{parthasarathy_controlled_2020}     
& Fixed length time series 'vehicle and engine speed' 
& DTW, SSIM\\
        \bottomrule
    \end{tabular}
\end{table}

\begin{table}[ht]
\caption{Experimental results comparing the performance of time series GANs for sinewave generation}
\begin{center}
\begin{tabular}{|c||c||c|c|c|}
    \hline
    \multirow{2}{*}{Architecture} &\multirow{2}{*}{Loss Function} &\multicolumn{3}{|c|}{Toy Sine Dataset}\\
    \cline{3-5}& & MMD &DTW &MSE\\
    \hline \hline
    \multirow{2}{*}{LSTM-LSTM} &BCE &-- &-- &--\\ 		
        &MSE &0.007805404	&54.16447988	&\textbf{0.148043822}\\
    \hline
    \multirow{2}{*}{BiLSTM-LSTM} &BCE &0.121597451	&428.43108	&3.070050828\\
        &MSE &0.951539381	&79.56070154	&0.236244962\\
    \hline
    \multirow{2}{*}{LSTM-CNN} &BCE &0.006319945	&55.36204204	&0.31545636\\
        &MSE &0.575763914	&86.73579748	&0.564357309\\
    \hline
    \multirow{2}{*}{BiLSTM-CNN} &BCE &\textbf{1.12958E-05}	&129.9257789	&0.919376137\\
        &MSE &0.489165394	&43.26942611	&0.186902029\\
    \hline
    \multirow{2}{*}{GRU-CNN} &BCE &0.024468314	&\textbf{37.1630491}	&0.230310881\\
        &MSE &0.372734849	&42.7348549	&0.228260543\\
    \hline
    \multirow{2}{*}{FC-CNN} &BCE &0.003933644	&58.35650673	&0.304803267\\
        &MSE &0.011720286	&43.36115622	&0.297288744\\
    \hline
\end{tabular}
\end{center}
\label{table:GAN-res-SINE}
\end{table}

\begin{table}[ht]
\caption{Experimental results comparing the performance of time series GANs for ECG generation on MIT-BIH Dataset}
\begin{center}
\begin{tabular}{|c||c||c|c|c|}
    \hline
    \multirow{2}{*}{Architecture} &\multirow{2}{*}{Loss Function} &\multicolumn{3}{|c|}{MIT-BIH Arrhythmia Dataset}\\
    \cline{3-5}& & MMD &DTW &MSE\\
    \hline \hline
    \multirow{2}{*}{LSTM-LSTM} &BCE &0.993149132	&30.18161173	&0.086768344\\ 		
        &MSE &0.8842204	&44.45535914	&0.138977587\\
    \hline
    \multirow{2}{*}{BiLSTM-LSTM} &BCE &0.991669978	&22.86345584	&0.069901973\\
        &MSE &0.973720273	&23.55338573	&0.080641843\\
    \hline
    \multirow{2}{*}{LSTM-CNN} &BCE &0.551958508	&\textbf{13.0158286}	&\textbf{0.015101118}\\
        &MSE &\textbf{0.0005273}	&24.73064185	&0.045702711\\
    \hline
    \multirow{2}{*}{BiLSTM-CNN} &BCE &0.924689511	&117.3994295	&0.227219626\\
        &MSE &0.068707491	&22.67402118	&0.058676694\\
    \hline
    \multirow{2}{*}{GRU-CNN} &BCE &0.005577049	&20.48459161	&0.033519309\\
        &MSE &0.770406707	&108.4124982	&0.194816291\\
    \hline
    \multirow{2}{*}{FC-CNN} &BCE &0.206880394	&23.99100792	&0.030977119\\
        &MSE &0.308249889	&18.23405152	&0.021292498\\
    \hline
\end{tabular}
\end{center}
\label{table:GAN-res-MITBIH}
\end{table}

\section{Privacy}
\label{sec:privacy}
As well as evaluating the quality of the data, a wide range of methods have been used to evaluate and mitigate the privacy risk associated with synthetic data created by GANs.

\subsection{Differential Privacy}
The goal of Differential Privacy is to preserve the underlying privacy of a database. An algorithm or, more specifically, a GAN achieves differential privacy if, by looking at the generated samples, we cannot identify whether the samples were included in the training set. As GANs attempts to model the training dataset, the problem of privacy lies in capturing and generating useful information about the training set population without the possibility of linkage from generated sample to an individuals data \cite{dwork_DP_2006}. 

As we have previously addressed, one of the main goals of GANs is to augment existing under-resourced datasets for use in further downstream applications such as upskilling of clinicians where healthcare data is involved. These personal sensitive data must contain privacy guarantees, and the rigorous mathematical definition of DP \cite{dwork_algorithmDP_2014} offers this assurance. 

Work is ongoing to develop machine learning methods with privacy-preserving mechanisms such as differential privacy (DP). Abadi \textit{et al.}  demonstrate the ability to train deep neural networks with DP and implement a mechanism for tracking privacy loss \cite{Abadi2016}. Xie \textit{et al.} proposed a differentially private GAN (DPGAN) that achieved differential privacy by adding noise gradients to the optimiser during the training phase \cite{Xie2018}. 

\subsection{Decentralised/Federated Learning}
Distributed or decentralised learning is another method for limiting the privacy risk associated with personal and personal sensitive data in machine learning. Standard approaches to machine learning require that all training data be kept on one server. Decentralised/distributed approaches to GAN algorithms require large communication bandwidth to ensure convergence \cite{augenstein2020generative, hardy2019mdgan} and are also subject to strict privacy constraints. A new method that enables communication efficient collaborative learning on a shared model while keeping all the training data decentralised is known as \textit{Federated learning} \cite{mcmahan2017Fedlearning}. Rasouli \textit{et al.} applied federated learning algorithm to a GAN for communication efficient distributed learning and proved the convergence of their federated learning GAN (FedGAN) \cite{rasouli2020fedgan}. However, it should be noted that they did not experiment with differential privacy in this study but note it as an avenue of future work. 

Combining the above techniques of federated learning and differential privacy in developing new GAN algorithms would lead to a fully decentralised private GAN capable of generating data without leakage of private information to the source data. This is clearly an open research avenue for the community.

\subsection{Assessment of Privacy Preservation}
We can also assess how well the generative model was able to protect our privacy through tests known as attribute and presence disclosure \cite{Choi2017}. The latter test is more commonly known in the machine learning space as a membership inference attack. This has become a quantitative assessment of how machine learning models leak information about the individual data records on which they were trained \cite{shokri2017membership}. 

Hayes \textit{et al.} carried out membership inference attacks on synthetic images and concluded that for acceptable levels of privacy in the GAN, the quality of the data generated is sacrificed \cite{Hayes2019}. Conversely, others have followed this approach and found that DP networks can successfully generate data that adheres to differential privacy and resists membership inference attacks without too much degradation in the quality of the generated data \cite{Esteban2017, Delaney2019, Brophy2020}.

\section{Discussion}
We have presented a survey of time series GAN-variants that have made significant progress in addressing the primary challenges identified in Section \ref{sec:challenges}. These GANs introduced the idea of both discrete and continuous sequential data generation and have made incremental improvements over one another via an architecture variant or a modified objective function capable of capturing the spatio-temporal dependencies present in these data types. The loss functions implemented in these works for some architectures will not necessarily generalise to others; hence they become architecture-specific. 
The architecture choices of the time series GANs affect both the quality and diversity of the data. However, there remain open problems in terms of the practical implementation of the generated data and GANs in real-world applications, particularly in health applications where the performance of these models can directly affect patients' quality of care/treatment.

The 'best' GAN architecture and objective function is still yet to be determined. As it stands, GANs tend to application-specific, that is, perform well for their intended purpose but do not generalise well beyond their original domain. A major limitation of time series GANs is the restrictions placed on the length of the sequence specified that the architecture can manage; documented experiments validating how well a time series GAN can adapt to varying data lengths are notably absent at the time of writing.

\section{Conclusion}
This paper reviews a niche but growing use of Generative Adversarial Networks for time series data based mainly around architectural evolution and loss function variants. We see that each GAN provides application-specific performance and doesn't necessarily generalise well to other applications, e.g. a GAN for generating high-quality physiological time series may not produce high-fidelity audio due to some limitation imposed by the architecture or loss function. A detailed review of the applications of time series GANs to real-world problems has been provided, along with their datasets and the evaluation metrics used for each domain. As stated in \cite{Wang2020}, GAN-related research for time series lags that of computer vision both in terms of performance and defined rules for generalisation of models. In conclusion, this review has highlighted the open challenges in this area and offers directions for future work and technological innovation, particularly for those GAN aspects related to evaluation, privacy, and decentralised learning. 

\section*{Acknowledgements}
This work is funded by Science Foundation Ireland under grant numbers 17/RC-PhD/3482 and SFI/12/RC/2289\_P2.

\bibliographystyle{IEEEtran}
\bibliography{template}

\end{document}